\documentclass{article}

    \usepackage[preprint]{neurips_2025}

\usepackage[utf8]{inputenc} 
\usepackage[T1]{fontenc}    
\usepackage{hyperref}       
\usepackage{url}            
\usepackage{booktabs}       
\usepackage{amsfonts}       
\usepackage{nicefrac}       
\usepackage{microtype}      
\usepackage{xcolor}         
\usepackage{amsmath} 
\usepackage{graphicx}
\usepackage{subcaption}
\usepackage{caption}
\usepackage{multirow}

\title{ACCORD: Autoregressive Constraint-satisfying Generation for COmbinatorial Optimization with Routing and Dynamic attention}

\author{
Henrik Abgaryan \\
LAMSADE, Université Paris Dauphine-PSL\\
Paris, France \\
\texttt{henrik.abgaryan@dauphine.eu}
\And
Tristan Cazenave\textsuperscript{*} \\
LAMSADE, Université Paris Dauphine-PSL\\
Paris, France \\
\texttt{tristan.cazenave@lamsade.dauphine.fr}
\And
Ararat Harutyunyan\textsuperscript{*} \\
LAMSADE, Université Paris Dauphine-PSL\\
Paris, France \\
\texttt{ararat.harutyunyan@lamsade.dauphine.fr}
}

\begin{document}

\maketitle

\renewcommand{\thefootnote}{\fnsymbol{footnote}}
\footnotetext[1]{Equal contribution.}

\begin{abstract}
Large Language Models (LLMs) have demonstrated impressive reasoning capabilities, yet their direct application to NP-hard combinatorial problems (CPs) remains underexplored. In this work, we systematically investigate the reasoning abilities of LLMs on a variety of NP-hard combinatorial optimization tasks and introduce \textbf{ACCORD}: \textbf{A}utoregressive \textbf{C}onstraint-satisfying generation for \textbf{CO}mbinatorial optimization with \textbf{R}outing and \textbf{D}ynamic attention. ACCORD features a novel dataset representation and model architecture that leverage the autoregressive nature of LLMs to dynamically enforce feasibility constraints, coupled with attention-based routing to activate problem-specific LoRA modules. We also present the ACCORD-90k supervised dataset, covering six NP-hard combinatorial problems: TSP, VRP, Knapsack, FlowShop, JSSP, and BinPacking. Extensive experiments demonstrate that our ACCORD model, built on an 8B-parameter Llama backbone, consistently outperforms standard prompting and input-output methods, even when compared to much larger LLMs, such as gpt-4. Ablation studies further show that our output structure enhances solution feasibility. To the best of our knowledge, this is the first large-scale, end-to-end framework for exploring the applications of LLMs to a broad spectrum of combinatorial optimization problems. The codes are publicly available at \footnote{https://github.com/starjob42/ACCORD}

\end{abstract}

\section{Introduction}

Large Language Models (LLMs) have rapidly established themselves as versatile engines for reasoning across a broad spectrum of tasks, encompassing arithmetic, commonsense logic , \cite{thoppilan2022lamda}, \cite{chowdhery2023palm}, \cite{brown2020language}. Among the prominent strategies enabling such capabilities is the Chain-of-Thought approach, which allows these models to decompose complex problems into sequential, interpretable steps \cite{wei2022chain}. 

Recent efforts have sought to adapt these reasoning techniques to address more advanced optimization tasks. Combinatorial optimization problems (CPs) are decision-making challenges where the goal is to select an optimal arrangement or subset from a large, discrete set of possibilities. Classic examples include the Traveling Salesman Problem (TSP), Vehicle Routing Problem (VRP), and Job Shop Scheduling Problem (JSSP), which have widespread applications in logistics, manufacturing, and artificial intelligence \cite{lenstra1979complexity}. Due to their NP-hard nature, even moderately sized instances possess a combinatorial explosion of potential solutions, rendering brute-force approaches infeasible. As a result, practical methods typically rely on heuristics or approximation algorithms to provide near-optimal solutions within reasonable time frames.
As NP-hard problems, CPs present huge obstacles in practical settings \cite{oroojlooyjadid2020applying}. Presently, the predominant paradigm in industry relies on metaheuristic algorithms—sophisticated combinations of simple, efficient heuristics—for solving CPs under various constraints. However, the success of these heuristics is often highly sensitive to the specific structure and requirements of each problem, necessitating tailored approaches for optimal results.

\begin{figure}[htbp] 
    \centering 
    \includegraphics[width=0.8\textwidth]{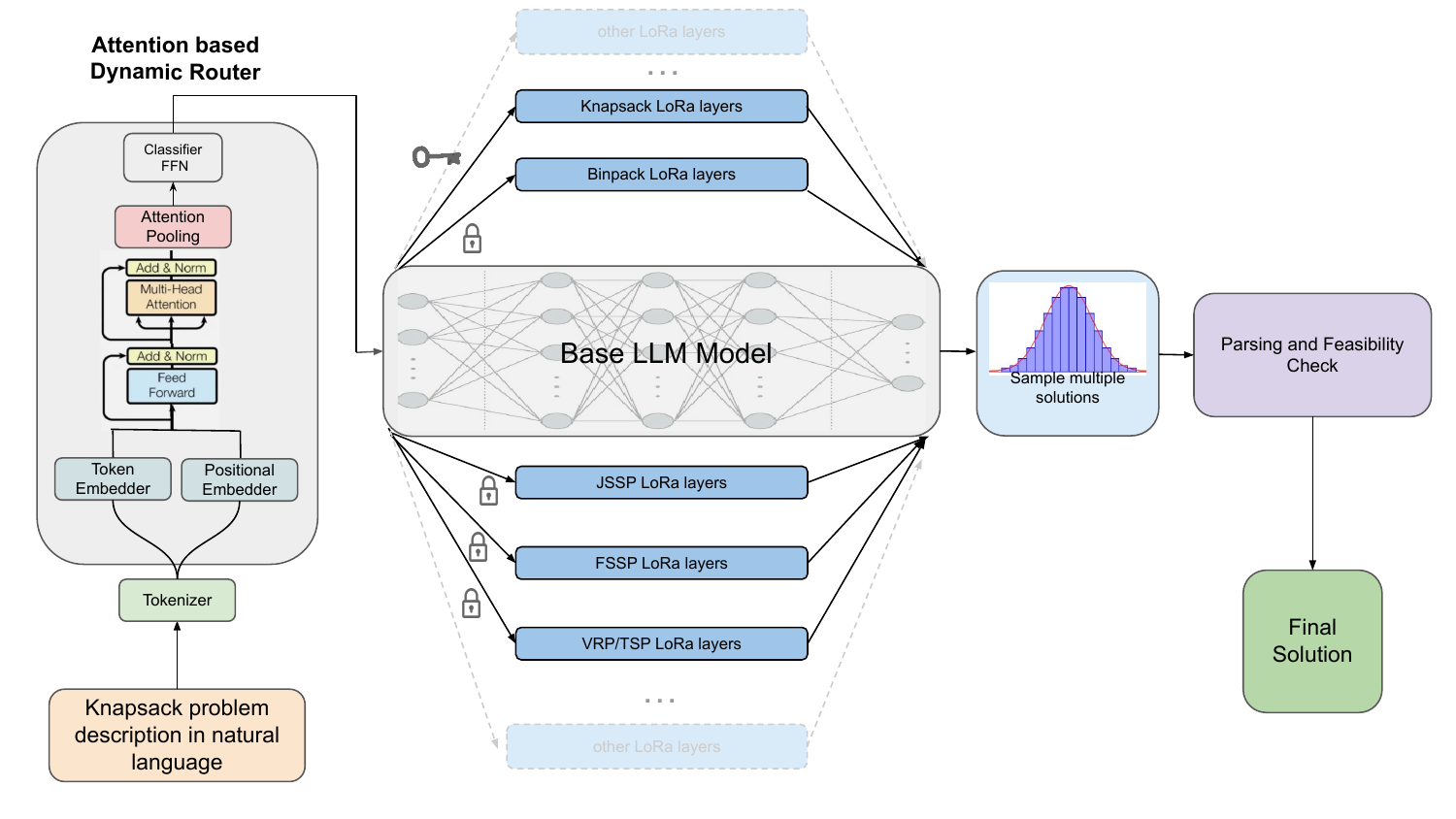}
    \caption{Overview of the ACCORD inference pipeline. As an example, a knapsack problem described in natural language is provided as input, then Attention based Dynamic router\ref{sec:model_arch} activates the corresponding LoRA layer specialized for knapsack tasks. Multiple candidate solutions are generated via sampling, each checked for feasibility. The best feasible solution is returned as the final output. Note that the pipeline generalizes to other combinatorial problems in the same manner; knapsack is shown here for illustration.}
    \label{fig:pipline} 
\end{figure}

At the same time, investigations into leveraging LLMs for combinatorial problem solving have revealed significant research gaps. While the latest breakthroughs highlight the promise of LLMs in diverse reasoning scenarios \cite{abgaryan2024starjob}, \cite{SGE}, \cite{wei2022emergent}, \cite{zhang2022automatic}, their full potential in the context of combinatorial optimization remains largely untapped.
Applying LLMs directly to these problems presents unique challenges: LLMs are trained primarily for natural language generation, not for enforcing strict combinatorial constraints, leading to issues such as hallucinations (plausible but infeasible solutions) \cite{huang2022improving}, lack of optimality, and limited interpretability \cite{valmeekam2022benchmark}. Furthermore, the absence of systematic search or explicit constraint mechanisms means LLM outputs can violate feasibility or fail to improve upon prior attempts. 
Recent advances have begun to explore the application of large language models (LLMs) to combinatorial optimization (CO). Numerous prompting-based approaches have been tested on CO tasks~\cite{yang2023optimization, huang2024large, Mittal2024, wei2022chain, zhou2022least, madaan2023self, SGE}, demonstrating progress in solution quality and constraint handling. However, to date, there has been no comprehensive study evaluating a unified fine-tuned LLM-based framework for NP-hard CO problems across multiple domains.

In this work, we address this gap by introducing \textbf{ACCORD} (\textbf{A}utoregressive \textbf{C}onstraint-satisfying generation for \textbf{CO}mbinatorial optimization with \textbf{R}outing and \textbf{D}ynamic attention), a novel framework for testing the reasoning capabilities of LLMs on combinatorial optimization problems. Our main contributions are as follows: (i) We propose the ACCORD-90k supervised dataset for TSP, VRP, Knapsack, FlowShop, JSSP, and BinPacking, employing an ACCORD representation that explicitly encodes problem constraints by leveraging the autoregressive nature of LLMs;(ii) we develop a model architecture that leverages attention-based dynamic routing and specialized Low-Rank Adaptation (LoRA) modules for different CO tasks; (iii) extensive ablation studies demonstrate that our method achieves lower optimality gaps and higher solution feasibility than both the traditional list-of-lists representation and state-of-the-art prompting techniques (including GPT-4 with the Code Interpreter enabled). Notably, it achieves substantial improvement percantage difference in feasibility over the list-of-list representation, with gains of 24.86\% in FlowShop, 7\% in JSSP, 4\% in Knapsack, and 2\% in BinPacking, 10\% in VRP and TSP, problems. To the best of our knowledge, this is the first work to demonstrate large-scale, end-to-end combinatorial problem solving with LLMs, offering new directions for testing symbolic reasoning and optimization within language models.

\section{Related Work}
\label{sec:related_work}

\subsection{Heuristic and Machine Learning Approaches on CO problems}
Combinatorial optimization has been tackled with both heuristic and exact methods. Simple priority dispatching rules (PDRs), such as shortest processing time or earliest due date, are computationally efficient but often yield suboptimal solutions due to their greedy nature~\cite{lenstra1979complexity}. Metaheuristics (e.g., simulated annealing, tabu search, genetic algorithms) offer improved solution quality, and exact approaches like the shifting bottleneck procedure~\cite{adams1988shifting}, mixed-integer programming, and constraint programming can find optimal solutions for small instances, though at high computational cost~\cite{roy1964machine, goel1996genetic}. Recently, machine learning, particularly deep reinforcement learning (RL) and graph neural networks (GNNs) have advanced combinatorial optimization~\cite{zhang2020learning,khalil2017learning,kool2019attention}. RL methods treat scheduling as sequential decision making, learning dispatching policies via environment interaction~\cite{zhang2020learning}. GNNs encode jobs and machines as nodes, enabling permutation-invariant representations and, when combined with RL, can model complex dependencies~\cite{khalil2017learning}. Attention-based and sequence-to-sequence models further enhance performance on tasks like TSP and VRP, often utilizing iterative refinement~\cite{kool2019attention}.

\subsection{Large Language Models in Combinatorial Optimization}

The advent of LLMs has introduced new paradigms for CO. Early work explored whether LLMs could generate solutions through prompting \cite{yang2023optimization}, \cite{huang2024large}, \cite{Mittal2024}, \cite{wei2022chain} \cite{zhou2022least}, \cite{madaan2023self}, \cite{SGE}. Prompting-based strategies, such as OPRO, involve iterative refinement based on feedback, while methods for VRP employ self-debugging and verification to enhance feasibility \cite{huang2024large}. However, scalability remains a challenge, as even strong prompting techniques struggle on larger or more complex instances \cite{Mittal2024}.  Recent research has explored a variety of prompting strategies to leverage LLMs for solving combinatorial optimization (CO) problems.
The \textbf{Input-Output (IO)} method presents the LLM with multiple examples of input and corresponding output solution pairs. The LLM is then prompted to generate an output solution in the same format as the provided examples. This approach relies on the LLM’s ability to generalize the mapping from input to output based on observed patterns. 
In \textbf{Chain-of-Thought (CoT)} prompting, the LLM is guided to produce a sequence of intermediate reasoning steps, or "thoughts,"  before arriving at the final answer \cite{wei2022chain}. This technique encourages the model to break down complex CO tasks into structured, stepwise reasoning, improving both transparency and solution quality. \textbf{Least-to-Most (LtM)} prompting strategy aims to decompose a complex problem into a sequence of simpler subproblems, solving them incrementally \cite{zhou2022least}. Each subproblem builds upon the solutions of previous ones, enabling the LLM to tackle challenging CO tasks through a series of manageable steps.  \textbf{Self-Refinement (SR)} is an iterative prompting technique wherein the LLM first generates an initial solution, then provides feedback on its own output, and finally refines the solution based on this feedback \cite{madaan2023self}. The process repeats until a satisfactory solution is reached. \textbf{Self-Guiding Exploration for Combinatorial Problems (SGE)} autonomously generates multiple thought trajectories for a given CO task \cite{SGE}. Each trajectory represents a distinct heuristic approach, inspired by metaheuristics. SGE decomposes these trajectories into actionable subtasks, executes them sequentially, and refines the results to ensure optimal solutions. 
Fine-tuning LLMs for CO tasks is another active area \cite{abgaryan2024starjob},\cite{Masoud2024} . \cite{abgaryan2024starjob} showed that fine-tuned LLM on job-shop scheduling, demonstrates significant improvements in solution quality. Similarly, \cite{Masoud2024} applied fine-tuning to TSP instances with promising but size-limited results. Hybrid methods integrate LLMs into evolutionary or search frameworks, where the LLM guides genetic operations or receives feedback from constraint solvers to iteratively improve solutions \cite{Liu2023, Wan2024, Awasthi2025}. While promising, these approaches often entail significant computational overhead and still face scaling hurdles.

\section{Preliminaries: Overview of Classic Combinatorial Optimization Problems}

In this section, we introduce several foundational combinatorial optimization problems, explaining their goals and constraints in accessible terms while also providing their standard mathematical formulations.

\begin{table}[!ht]
\centering
\caption{Optimality gap (\%) of prompting methods (GPT-4 with code interpreter) vs. ACCORD (Llama 8B). Lower is better. N/A: no feasible solution.}
\label{tab:com_with_promp}
\scriptsize
\begin{tabular}{|c|c|ccccc|}
\hline
Size & Method & Knapsack & BinPack & TSP & VRP & JSSP \\
\hline
\multirow{7}{*}{5} 
& IO (GPT-4)      & 90.1 & 108.2 & 100.3 & 102.0 & 105.3 \\
& CoT (GPT-4)     & 66.9 & 78.2  & 81.2  & 78.2  & 79.4 \\
& SR (GPT-4)      & 62.0 & 77.4  & 71.6  & 72.5  & 71.7 \\
& LtM (GPT-4)     & 21.6 & 40.0  & 43.6  & 40.7  & 44.1 \\
& SGE (GPT-4)     & 8.1  & 9.1   & 8.3   & 11.9  & 9.3  \\
& IO (Llama 8B)      & N/A  & N/A   & N/A   & N/A   & N/A  \\
& ACCORD (Llama 8B)  & \textbf{3.9*} & \textbf{0.0*} & \textbf{0.6*} & \textbf{1.0*} & \textbf{0.0*} \\
\hline
\multirow{7}{*}{8} 
& IO (GPT-4)      & 103.5 & 112.8 & 116.9 & 116.3 & 108.2 \\
& CoT (GPT-4)     & 73.8  & 85.1  & 89.0  & 89.5  & 85.2  \\
& SR (GPT-4)      & 72.6  & 86.3  & 85.6  & 83.3  & 78.4  \\
& LtM (GPT-4)     & 26.4  & 52.7  & 53.5  & 54.4  & 49.8  \\
& SGE (GPT-4)     & 14.9  & 21.0  & 15.2  & 19.7  & 21.3  \\
& IO (Llama 8B)      & N/A   & N/A   & N/A   & N/A   & N/A   \\
& ACCORD (Llama 8B)  & \textbf{7.4*} & \textbf{0.0*} & \textbf{1.8*} & \textbf{1.0*} & \textbf{5.0*} \\
\hline
\multirow{7}{*}{12}
& IO (GPT-4)      & 101.5 & 120.7 & 121.6 & 118.5 & 117.6 \\
& CoT (GPT-4)     & 79.3  & 93.8  & 86.8  & 90.1  & 89.3  \\
& SR (GPT-4)      & 77.1  & 82.2  & 88.6  & 88.4  & 87.0  \\
& LtM (GPT-4)     & 35.8  & 55.4  & 57.5  & 59.2  & 56.0  \\
& SGE (GPT-4)     & 16.8  & 22.4  & 16.1  & 24.0  & 22.9  \\
& IO (Llama 8B)      & N/A   & N/A   & N/A   & N/A   & N/A   \\
& ACCORD (Llama 8B)  & \textbf{5.1*} & \textbf{2.6*} & \textbf{2.9*} & \textbf{2.2*} & \textbf{12.4*} \\
\hline
\end{tabular}
\end{table}

\textbf{General Combinatorial Optimization Problem}
Combinatorial optimization involves searching for the best solution from a finite set of possibilities. Formally, given a set of feasible solutions $\mathcal{S}$ and an objective function $f: \mathcal{S} \rightarrow \mathbb{R}$, the goal is to find
\[
s^* = \arg\min_{s \in \mathcal{S}} f(s)
\]
or, in some cases, to maximize $f(s)$ depending on the problem.

\textbf{Traveling Salesman Problem (TSP)}
Given a list of cities and the distances between each pair of cities, what is the shortest possible route that visits each city exactly once and returns to the starting point. 
  Mathematically, for $n$ cities $V = \{1, 2, \dots, n\}$ and a distance matrix $D \in \mathbb{R}^{n \times n}$, we seek a tour (a permutation $\pi$ of all cities) that minimizes the total travel distance, where $\pi(n+1) = \pi(1)$ to ensure the tour closes:
\[
\min_{\pi \in \mathcal{P}_n} \sum_{i=1}^{n} D_{\pi(i), \pi(i+1)}
\]

\textbf{Vehicle Routing Problem (VRP)}  
The VRP extends the TSP to multiple vehicles. Given a depot, $n$ customers (with demands $q_i$), and a fleet of vehicles each with capacity $Q$, the goal is to design routes—each starting and ending at the depot—so that every customer is visited exactly once, no vehicle exceeds its capacity, and the total travel distance is minimized:
\[
\min \sum_{k=1}^m \sum_{j=0}^{\ell_k} D_{v^k_j, v^k_{j+1}}
\]
subject to
\begin{align*}
& \bigcup_{k=1}^m \{v^k_1, \dots, v^k_{\ell_k}\} = V \quad \text{(All customers served)} \\
& \sum_{j=1}^{\ell_k} q_{v^k_j} \leq Q \quad \forall k \quad \text{(Capacity constraint)}
\end{align*}

\textbf{Job Shop Scheduling Problem (JSSP)}  
JSSP schedules $n$ jobs, each as a sequence of operations on specific machines. Each operation $O_{j,k}$ requires machine $M_{j,k}$ for $p_{j,k}$ time units, following job order. Let $S_{j,k}$ and $C_{j,k}$ be the start and completion times. The objective is to minimize makespan:
\[
\min C_{\max} = \max_{j} C_{j, \ell_j}
\]
subject to:
\begin{align*}
& \text{(Precedence)} \quad S_{j,k+1} \geq C_{j,k} \\
& \text{(No machine conflicts)} \quad S_{j,k} \geq C_{j',k'} \text{ or } S_{j',k'} \geq C_{j,k}, \\
&\qquad \forall (j,k) \neq (j',k') \text{ with } M_{j,k} = M_{j',k'}
\end{align*}
\textbf{Knapsack Problem (KP)}
Given a set of items, each with a value and weight, what is the most valuable combination of items you can carry without exceeding the weight limit of your knapsack. With $n$ items (weights $w_i$, values $v_i$) and capacity $W$, choose $x_i \in \{0,1\}$ (item picked or not) to solve:
\[
\max \sum_{i=1}^n v_i x_i \qquad \text{s.t.} \qquad \sum_{i=1}^n w_i x_i \leq W
\]

\textbf{Bin Packing Problem (BPP)}
Given a set of items of varying sizes, how can you pack them into the fewest number of fixed-size bins. For $n$ items of sizes $s_i \in (0,1]$, assign them to bins of capacity $1$ so as to minimize the total number of bins $K$:
\[
\min K
\]
subject to:
\[
\sum_{i \in B_k} s_i \leq 1 \quad \forall k, \qquad \bigcup_{k=1}^K B_k = \{1, \dots, n\}, \qquad B_k \cap B_{k'} = \emptyset \ \forall k \neq k'
\]
where $B_k$ is the set of items in bin $k$.

\textbf{Flow Shop Scheduling Problem (FSSP)}

We consider the permutation flow shop scheduling problem (FSSP), where all jobs are processed in the same order on each of $m$ machines. Given $n$ jobs $J = \{1, 2, \ldots, n\}$ and processing times $p_{j,k}$ for job $j$ on machine $k$, the goal is to find a job sequence $\pi$ that minimizes the makespan.

Let $C_{j,k}$ denote the completion time of job $j$ on machine $k$. The constraints are:
\begin{align*}
& \text{(Machine order)} \quad C_{\pi(i),k} \geq C_{\pi(i),k-1} + p_{\pi(i),k} \quad \forall i=1,\ldots,n,\, k=2,\ldots,m \\
& \text{(Job sequence)} \quad C_{\pi(i),k} \geq C_{\pi(i-1),k} + p_{\pi(i),k} \quad \forall i=2,\ldots,n,\, k=1,\ldots,m
\end{align*}

The objective is to find the permutation $\pi^*$ that minimizes the makespan:
\[
\min_{\pi \in \mathcal{P}_n} C_{\max} = \min_{\pi \in \mathcal{P}_n} \max_{i} C_{\pi(i), m}
\]
where $C_{\pi(i), m}$ is the completion time of job $\pi(i)$ on the last machine.

\section{Main Method: ACCORD Representation for Feasibility-Aware Solution Generation}
\label{sec:accord}

A core challenge in applying Large Language Models (LLMs) to combinatorial optimization is the effective encoding of feasibility constraints within the generated solutions. Conventional representations, such as the ``list of lists'' format, provide direct encodings of solution sets, which are familiar to LLMs due to their prevalence in general-purpose data and code corpora. However, these representations are static—constraints are only checked after solution generation, offering limited guidance for incremental feasibility during the autoregressive decoding process. To address this limitation, we decided to utilize the auto-regressive nature of the LLMs and developed a representation, which is specifically designed to leverage the autoregressive generation paradigm of LLMs. Unlike the list-based format, our representation decomposes solutions into a sequence of state transitions, with each step not only specifying the next element of the solution but also explicitly updating and exposing the relevant feasibility metrics (e.g., cumulative weights, distances, machine usage, or value). This design allows the model to compute and check constraints dynamically as each token is generated, closely mimicking the typical reasoning and verification process of a human solver.
ACCORD representation embeds constraint satisfaction directly into the generation process. For instance, in the Knapsack problem, each item addition is accompanied by an explicit update of the running total value and weight, immediately verifying the capacity constraint at each step:
\begin{quote}
    \texttt{[[item\_id, weight, value] -> value: prev\_v + value = new\_v, weight: prev\_w + weight = new\_w <= capacity], ...}
\end{quote}
Please refer to page 1 in Appendix \ref{app:example_solution} for a concrete example. Similarly, for Bin Packing, the incremental assignment of items to bins is annotated with cumulative weights, ensuring that no bin exceeds its capacity as the sequence unfolds. Routing problems (VRP, TSP) and scheduling problems (JSSP) are analogously handled by tracking cumulative distances or machine times within the autoregressive output stream. Example of each of these generates is avaialable in the Appendix \ref{app:example_solution}. This approach transforms the constraint satisfaction problem into a stepwise process, where feasibility checks are interleaved with generation. As a result, the LLM is naturally guided away from infeasible sequences, as each decision is immediately contextualized by the current state of the solution.

\subsection{Dataset Generation}
\label{sec:dataset}

We generated synthetic supervised datasets for several CO problems using Google OR-Tools \cite{ortools} as the solver. For each instance, solutions were produced in both the conventional ``list of lists'' and ACCORD representations. Roughly 15,000 instances were created per problem type.
\textbf{TSP \& VRP:} Instances varied in location count ($N \in \{5, 8, \ldots, 100\}$) and number of vehicles ($V \in \{1, \ldots, 10\}$), with random coordinates and demands. OR-Tools solved these using the `PATH\_CHEAPEST\_ARC` strategy. \textbf{Knapsack:} Item counts ($N \in \{5, \ldots, 100\}$) and difficulty were varied, influencing item properties and constraints. Optimal solutions were computed using OR-Tools, discarding instances that timed out. \textbf{Bin Packing:} Instances varied by item count, weight limits, and target bin numbers. Bin capacities were set accordingly, and OR-Tools was used to minimize bin usage under a timeout. \textbf{JSSP:} Job Shop Scheduling instances ranged from $10 \times 10$ to $100 \times 20$ jobs and machines, with random operation sequences and durations. Solutions minimized makespan using the CP-SAT solver. \textbf{FSSP:} Permutation Flowshop instances ranged from $5 \times 1$ to $50 \times 2$ and $2 \times 50$, with random processing times. Solutions were generated with the NEH heuristic~\cite{nawaz1983heuristic}. More details avalable in Appendix Section \ref{sec:dataset_appendix}.

\begin{figure}[htbp]
    \centering
    \begin{subfigure}[t]{0.45\textwidth}
        \centering
        \includegraphics[width=\linewidth]{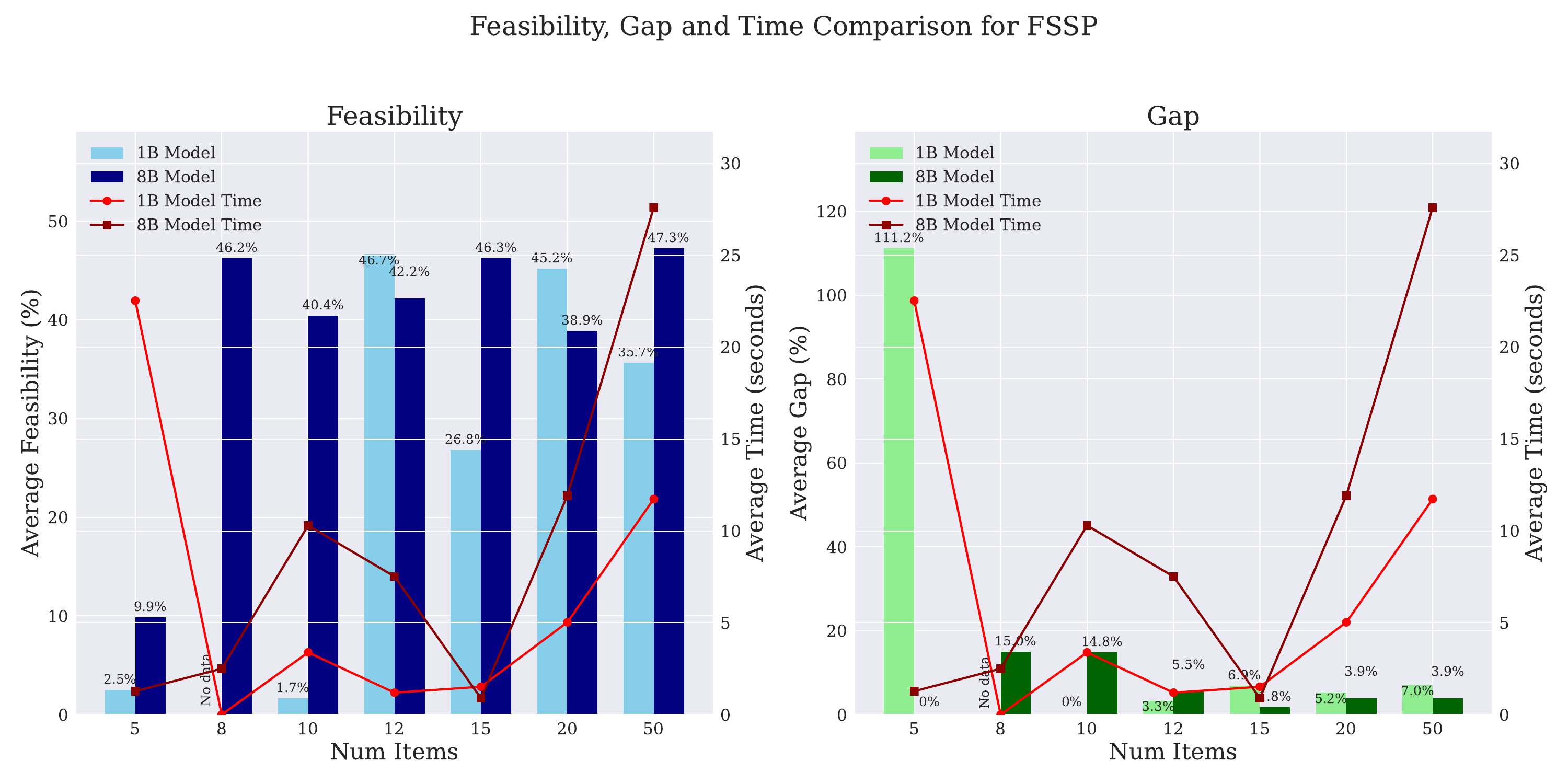}
        \caption{Flow Shop}
    \end{subfigure}\hfill
    \begin{subfigure}[t]{0.45\textwidth}
        \centering
        \includegraphics[width=\linewidth]{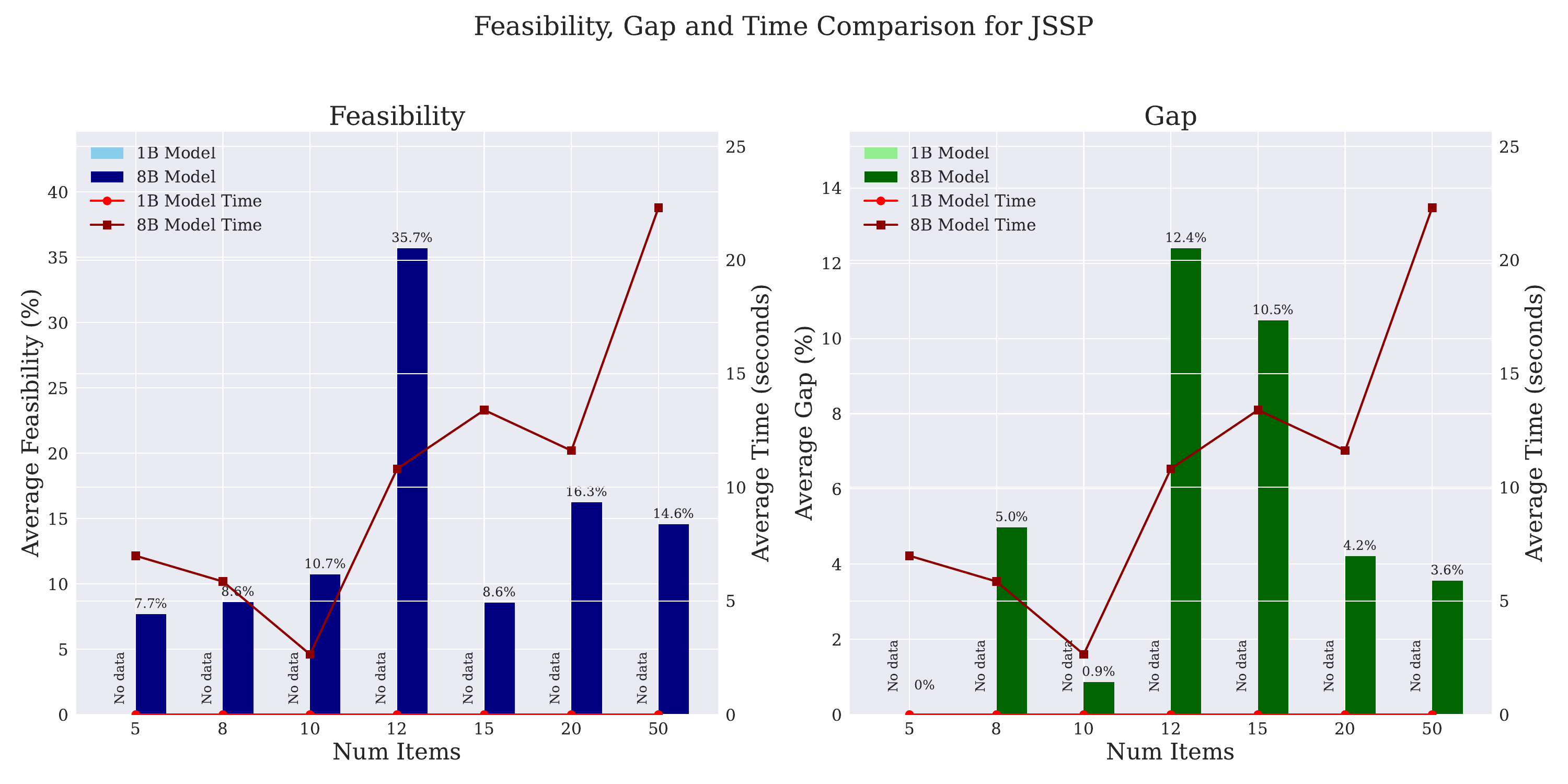}
        \caption{JSSP}
    \end{subfigure}

    \vspace{1em} 

    \begin{subfigure}[t]{0.45\textwidth}
        \centering
        \includegraphics[width=\linewidth]{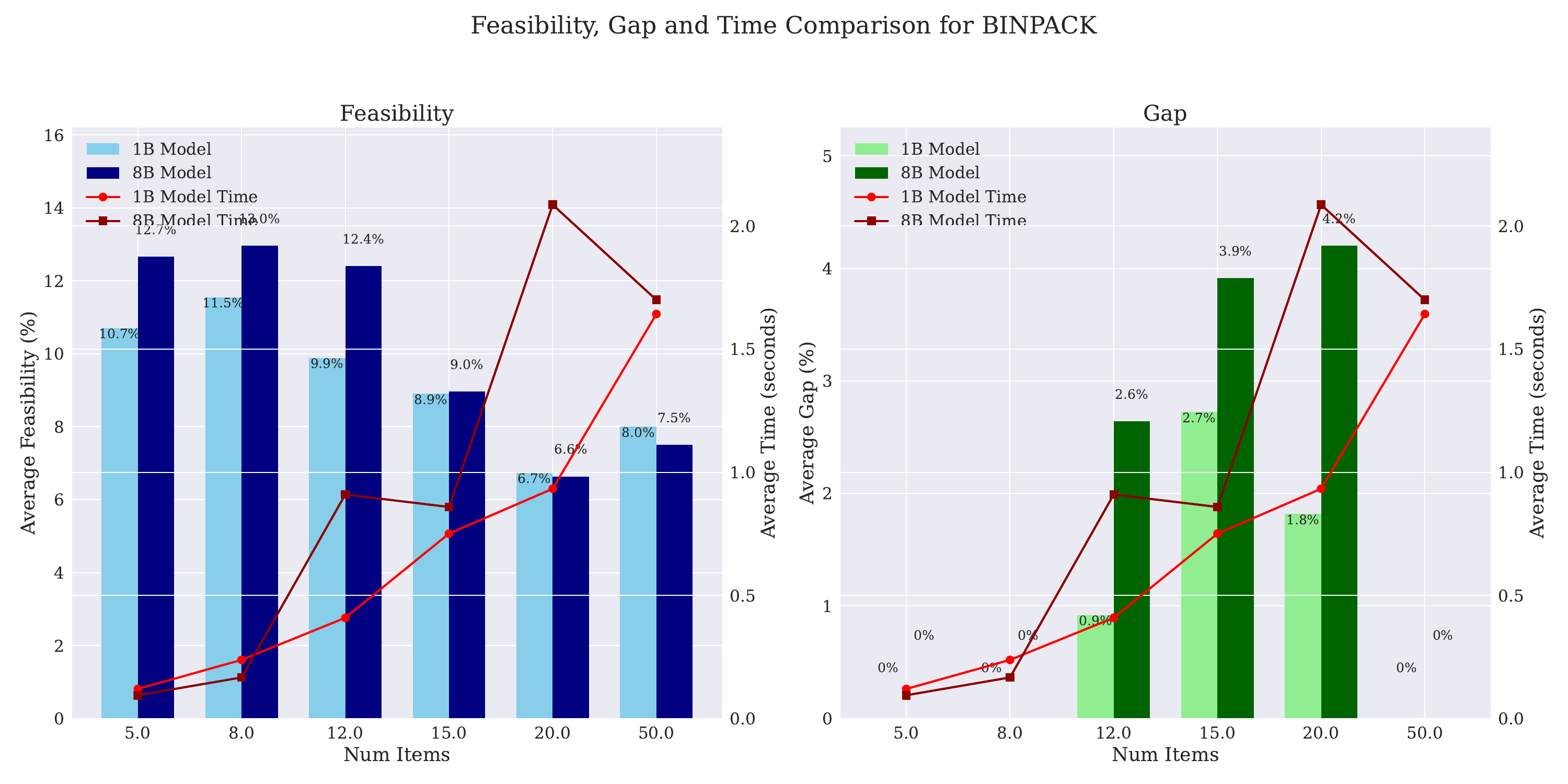}
        \caption{BinPack}
    \end{subfigure}\hfill
    \begin{subfigure}[t]{0.45\textwidth}
        \centering
        \includegraphics[width=\linewidth]{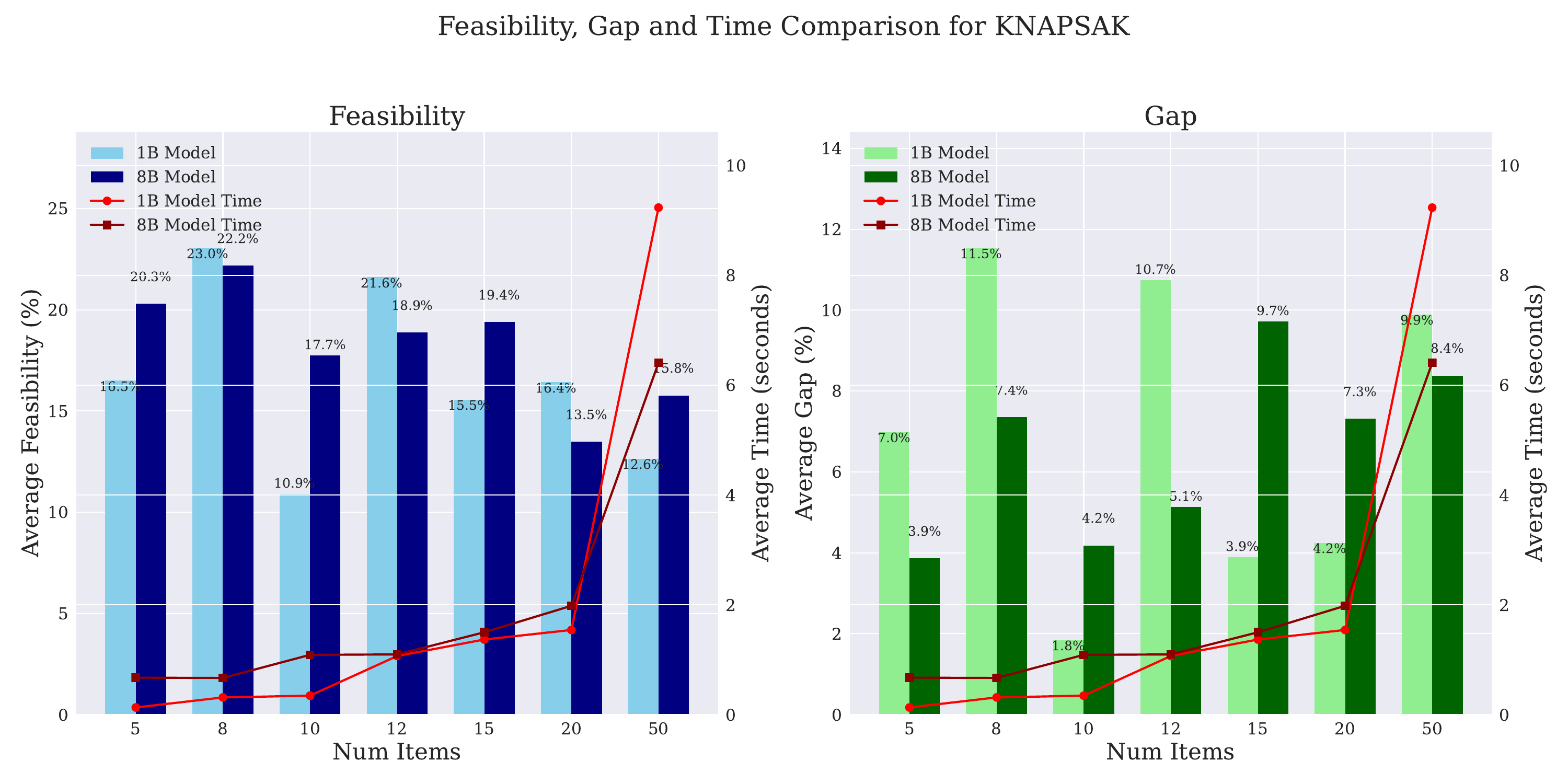}
        \caption{Knapsack}
    \end{subfigure}

    \vspace{1em} 

    \begin{subfigure}[t]{0.45\textwidth}
        \centering
        \includegraphics[width=\linewidth]{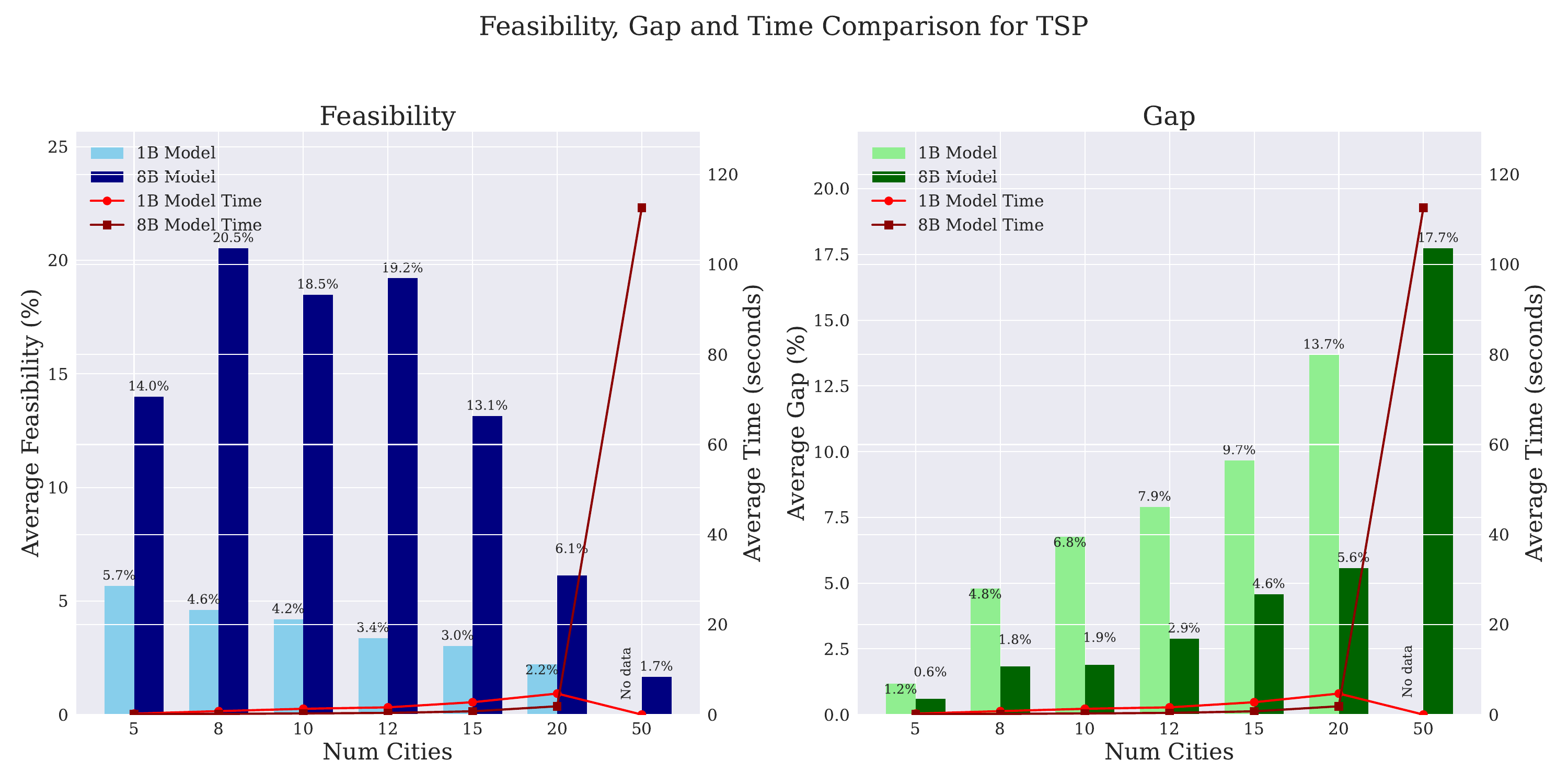}
        \caption{TSP}
    \end{subfigure}\hfill
    \begin{subfigure}[t]{0.45\textwidth}
        \centering
        \includegraphics[width=\linewidth]{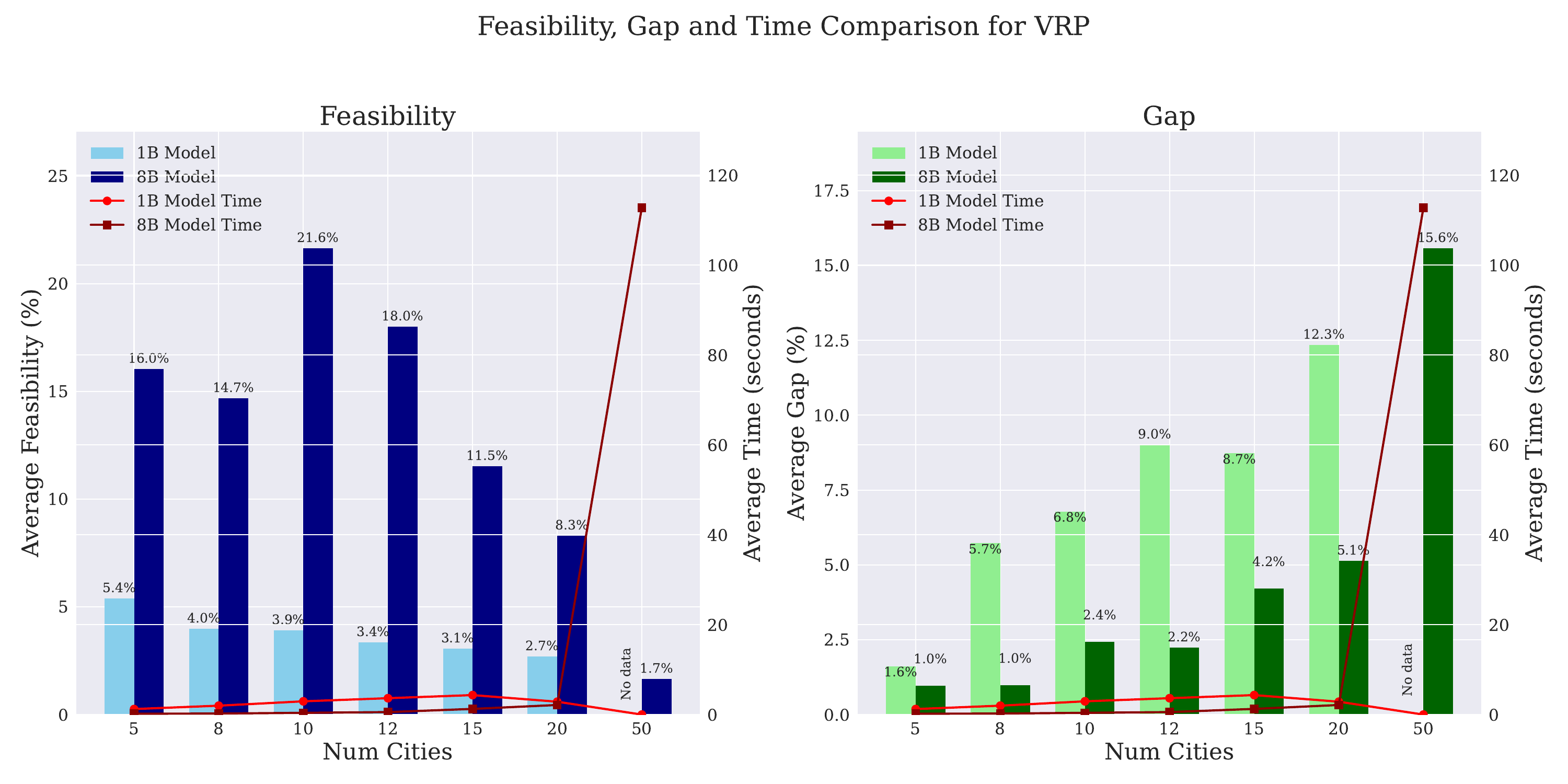}
        \caption{VRP }
    \end{subfigure}
    \caption{This figure illustrates the performance of the LLama 3.1 (8B) and LLama 3.2 (1B) models in terms of the average gap percentage compared to the OR-Tools solution, where a lower gap indicates better performance. The left y-axis represents the average gap percentage, while the right y-axis corresponds to the running time in seconds. Bar plots indicate the average gap. The line plots depict the average running time per instance size, with the x-axis showing the problem size in terms of the number of nodes in the graph representation.  Instances labeled as "No Data" indicate that, within a sampling budget of 60, the model failed to generate any feasible solution.}
    \label{fig:model_size_effect}
\end{figure}

\section{Model Architecture}
\label{sec:model_arch}
To dynamically activate the correct LoRA layers for each combinatorial optimization problem, we use an attention-based Dynamic Router TextClassifier that selects the appropriate LoRA weights based on the instruction text (see Figure~\ref{fig:pipline}). Our model builds on a transformer architecture, enhanced to capture problem-specific features. Each input token $x_i$ is embedded with positional information and normalized:

\begin{equation}
    \mathbf{E}' = \text{Dropout}(\text{LayerNorm}(\mathbf{E}_{\text{token}}(\mathbf{x}) + \mathbf{E}_{\text{pos}}(\mathbf{p})))
\end{equation}

The embeddings are projected to the hidden dimension and passed through several transformer layers with alternating multi-head attention and feed-forward sublayers, each followed by layer normalization. Token representations from the final transformer layer are pooled using attention-based pooling:

\begin{align}
    \mathbf{r} &= \sum_{i=1}^{n} a_i \mathbf{h}_i \qquad
    \mathbf{y} = \mathbf{W}_2 \cdot \text{LayerNorm}(\text{GELU}(\mathbf{W}_1 \mathbf{r} + \mathbf{b}_1)) + \mathbf{b}_2
\end{align}

Finally, the pooled vector $\mathbf{r}$ is passed through a classification head to produce logits $\mathbf{y}$ for each problem class. This architecture enables dynamic, instruction-based activation of problem-specific LoRA adapters.

\section{Training Details}
\label{sec:train_details}
We conducted supervised fine-tuning using input-output pairs for two models from Meta: Llama 3.1 8B and Llama 3.2 1B. To minimize memory usage during training, we employed 4-bit quantized versions of these models and trained each for 2 epochs. For a fair comparison, we fine-tuned each model with the same hyperparameters, varying only the output representation: once using the list-of-lists format and once using the ACCORD format, while keeping the input and all other hyperparameters identical. We used Rank-Stabilized Low-Rank Adaptation (RSLoRA) \cite{kalajdzievski2023rankstabilizationscalingfactor} with a rank of \(r = 64\) and \(\alpha = 64\). The two epochs, training required roughly 40 hours and about 30GB of GPU memory on Nvdidia RTX A6000 GPU. We limited the context length of the model to 40k instead of the original 128k, to reduce memory consumption and increase the speed of fine-tuning. ``Context length''
refers to the maximum number of tokens (words or subwords) the model can process at once as input. More training details and curves are available in \ref{app:train_details} and in Figure \ref{fig:loss_plots}.

\begin{figure}[htbp]
    \centering
    \begin{subfigure}[t]{0.4\textwidth}
        \centering
        \includegraphics[width=\linewidth]{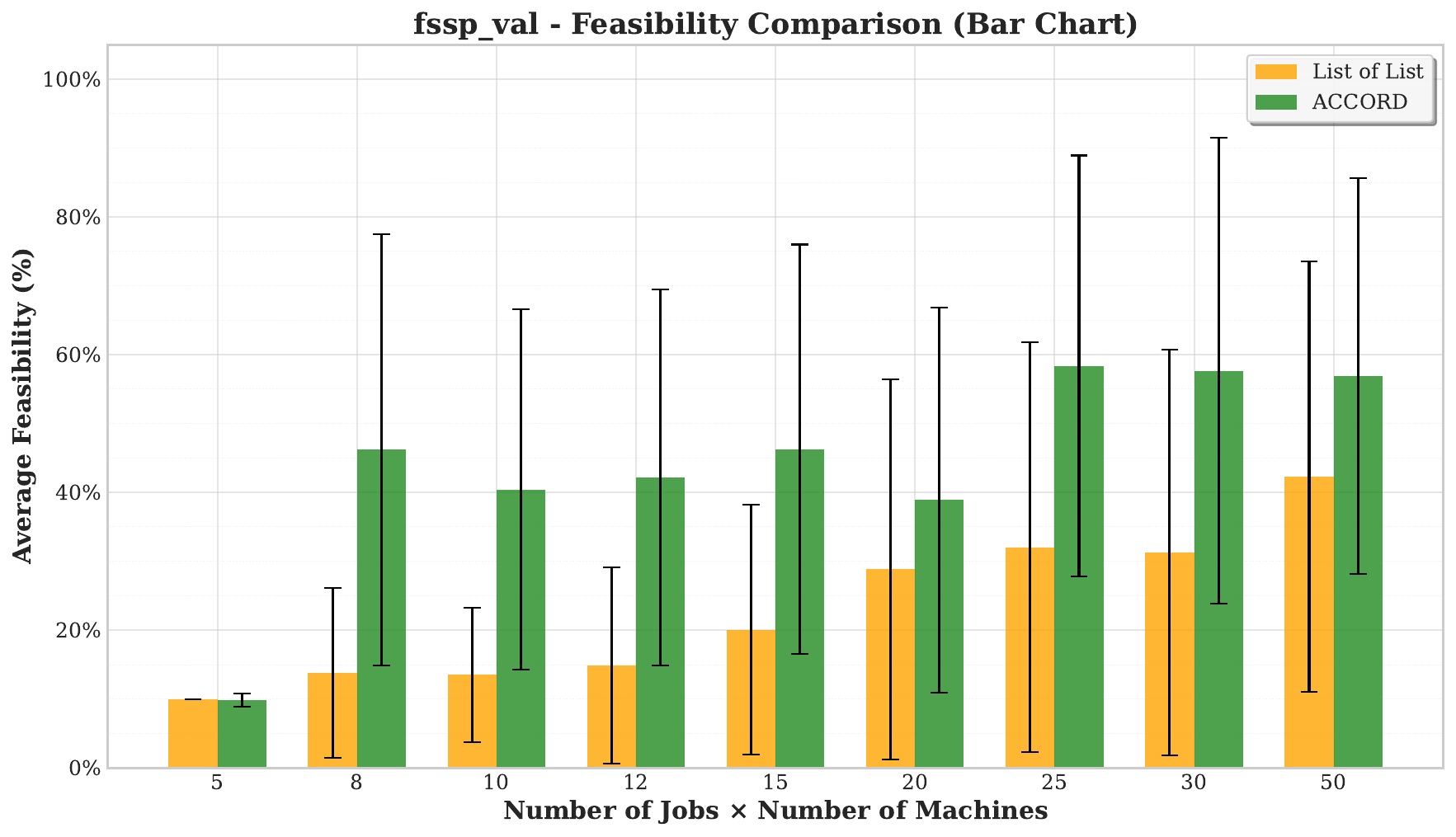}
        \caption{FSSP}
    \end{subfigure}\hfill
    \begin{subfigure}[t]{0.4\textwidth}
        \centering
        \includegraphics[width=\linewidth]{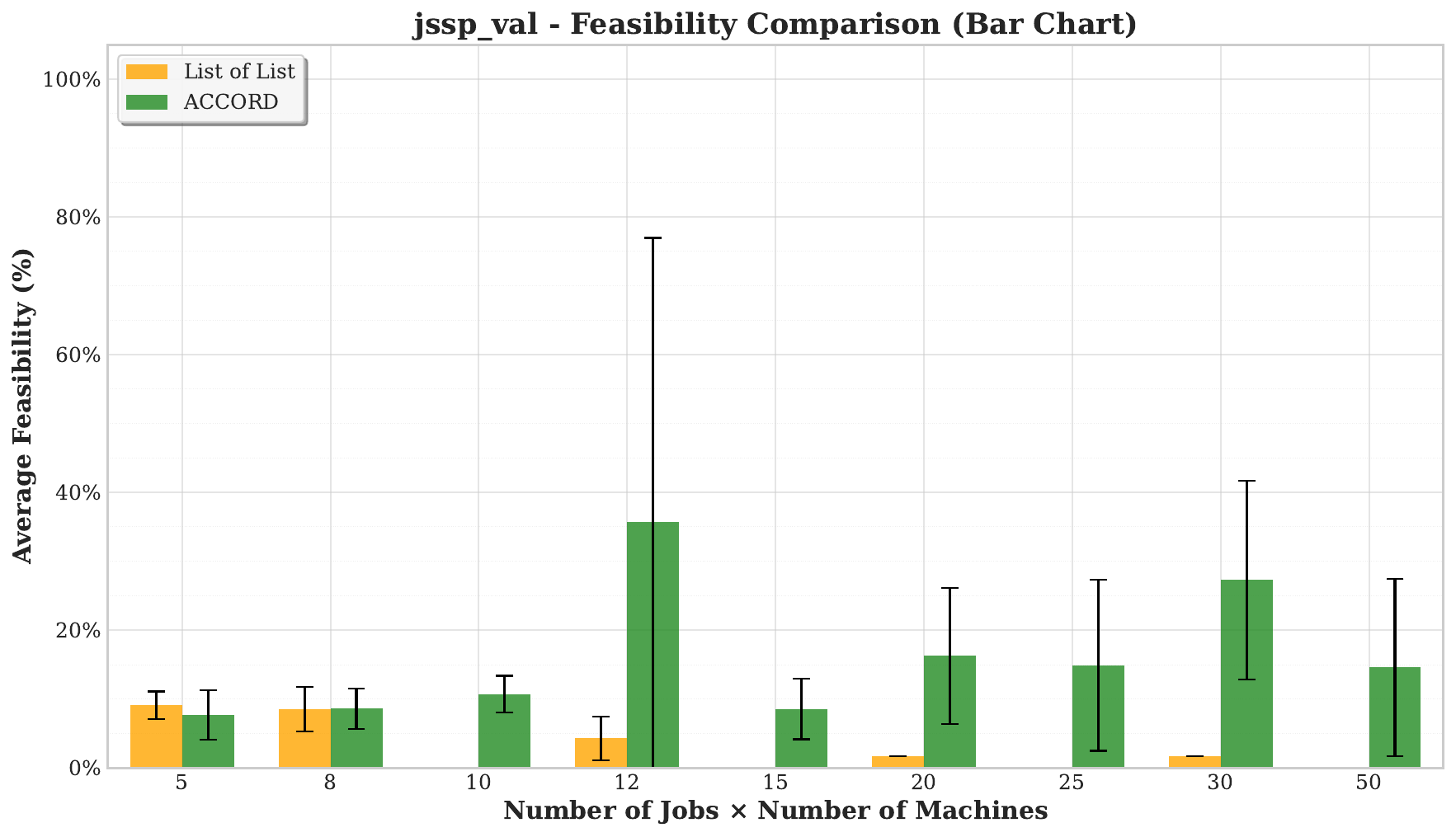}
        \caption{JSSP}
    \end{subfigure}

    \vspace{0.2em} 

    \begin{subfigure}[t]{0.4\textwidth}
        \centering
        \includegraphics[width=\linewidth]{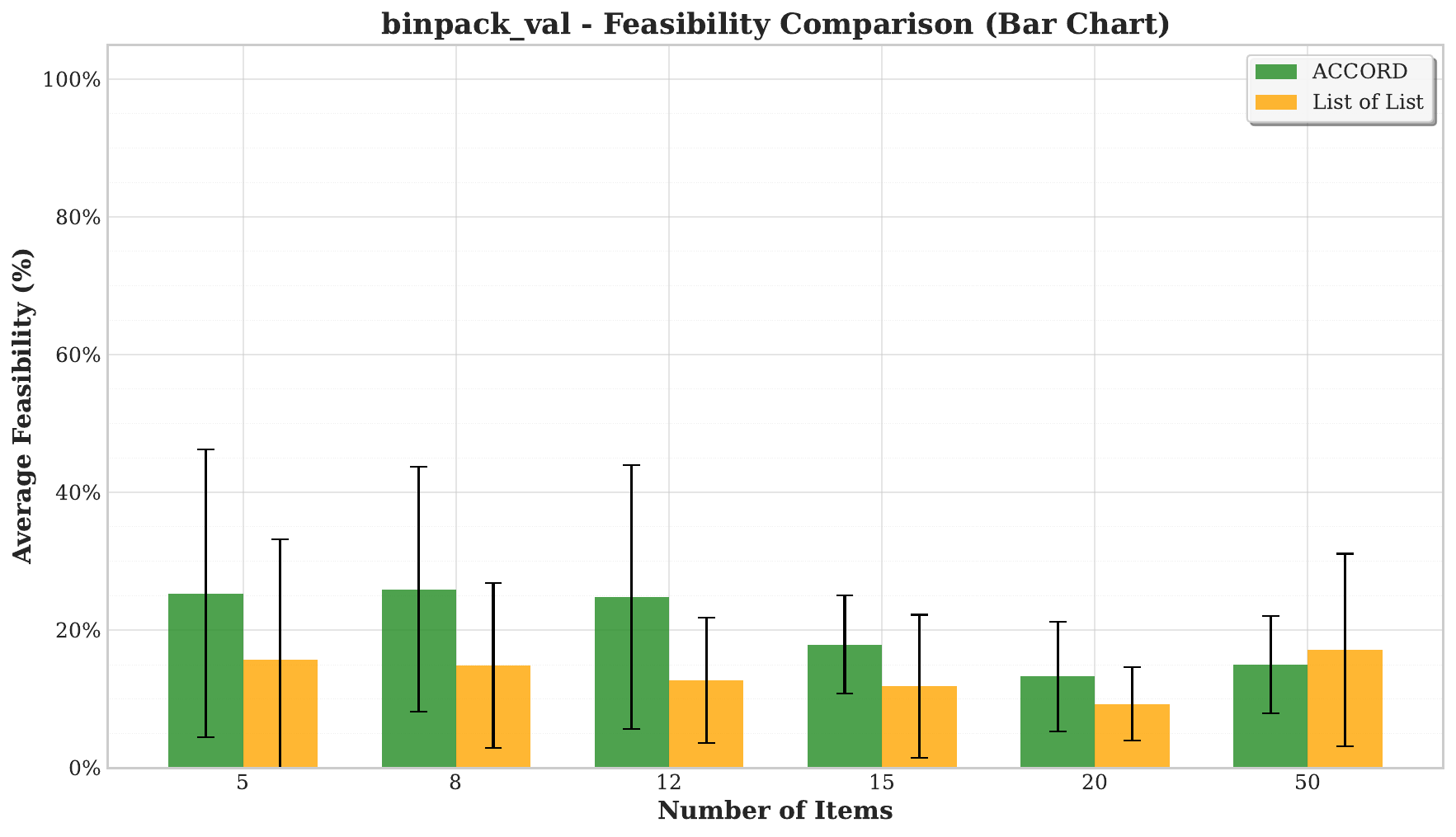}
        \caption{Bin Packing}
    \end{subfigure}\hfill
    \begin{subfigure}[t]{0.4\textwidth}
        \centering
        \includegraphics[width=\linewidth]{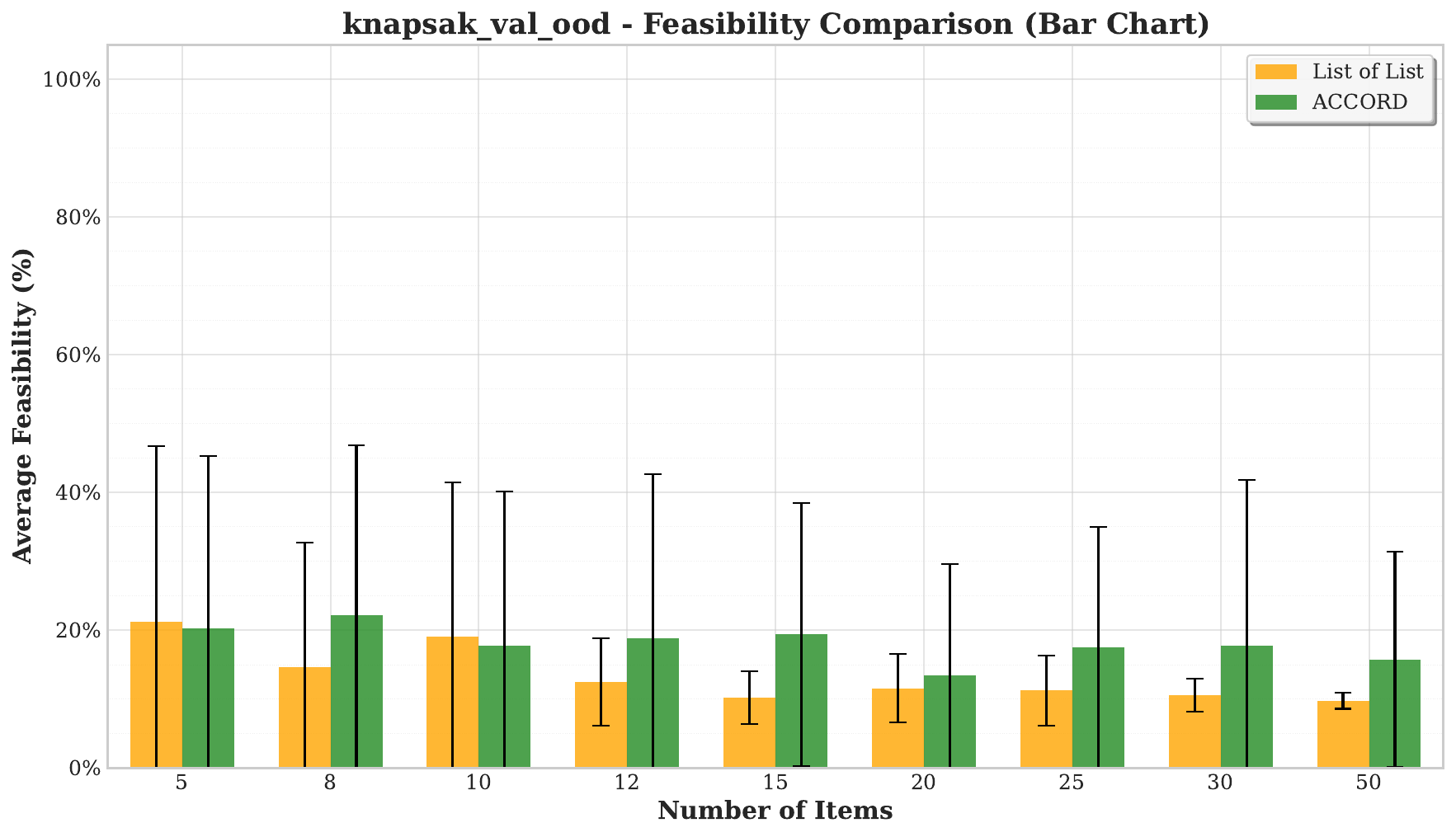}
        \caption{Knapsack}
    \end{subfigure}

    \vspace{0.2em} 

    \begin{subfigure}[t]{0.4\textwidth}
        \centering
        \includegraphics[width=\linewidth]{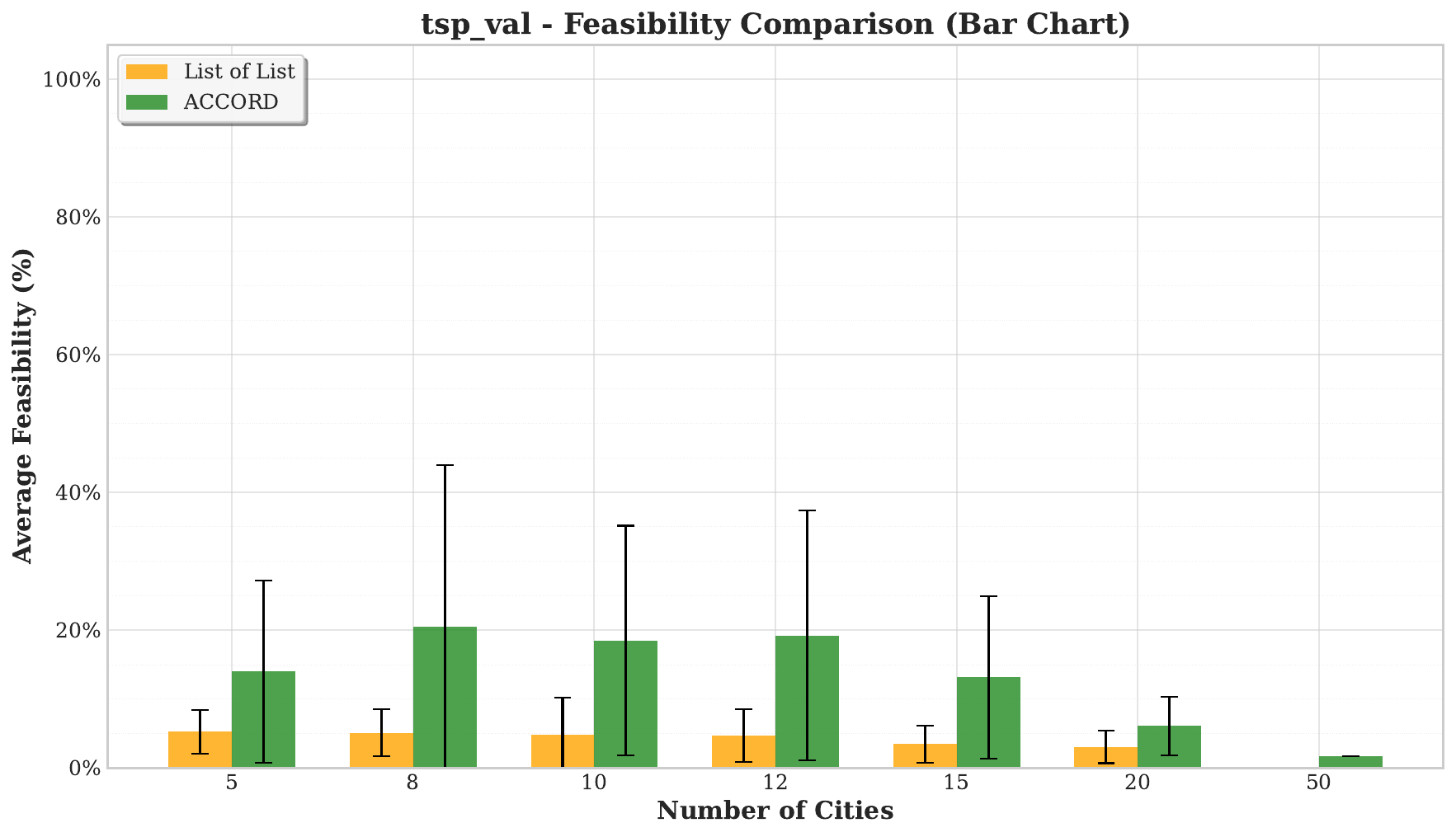}
        \caption{TSP}
    \end{subfigure}\hfill
    \begin{subfigure}[t]{0.4\textwidth}
        \centering
        \includegraphics[width=\linewidth]{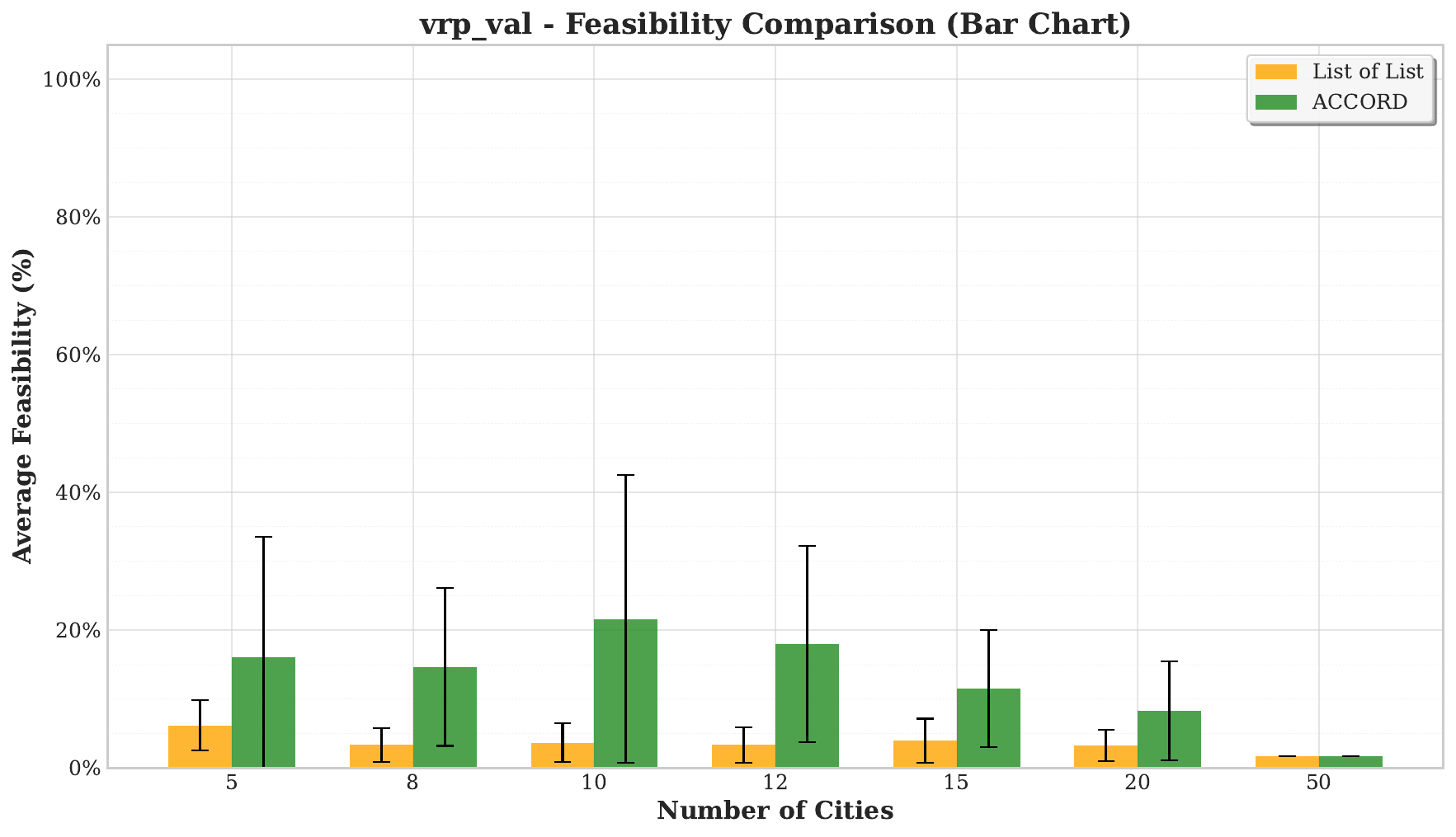}
        \caption{VRP}
    \end{subfigure}
    \caption{Average feasibility comparison with OR-Tools solution across different problem instance sizes; the higher the feasibility percentage, the better.}
    \label{fig:avg_feasibility_comparison}
\end{figure}

\subsection{Empirical Comparison with List-of-List Representation}
\label{sec:emp_comparison}

We empirically evaluate the impact of problem representation by fine-tuning LLama 3.1 8B on both list-of-list and ACCORD formats with identical hyperparameters and input (see Section~\ref{sec:dataset}), using a validation set of 100 out-of-distribution instances for each problem size ($n \in \{5, 8, 10, 12, 15, 20, 25, 30, 50\}$). The inference pipeline (Fig.~\ref{fig:pipline}) employs an Attention-Based Dynamic Router (Section~\ref{sec:model_arch}) to select the appropriate LoRA branch, generating 60 candidate solutions per instance. Each solution is checked for feasibility, and the best feasible solution, i.e., the one with the lowest optimality gap is selected as the final output. The optimality gap is defined as
\begin{equation}
    \text{Gap} = \frac{\text{Model Value} - \text{OR-Tools Value}}{\text{OR-Tools Value}},
\end{equation}
where a lower gap indicates a better solution. Feasibility is measured as the percentage of generated solutions that satisfy all constraints. Our results (Fig.~\ref{fig:avg_feasibility_comparison}) show that, although list-of-list representation is familier to LLMs, models trained with this format tend to ignore feasibility constraints, resulting in lower feasibility rates and higher optimality gaps. In contrast, the ACCORD representation explicitly encodes feasibility into the output, enabling the LLM to produce a larger proportion of valid and near-optimal solutions, particularly as the problem size increases. Table~\ref{tab:com_with_promp} further compares our method against various prompting strategies (see Section~\ref{sec:related_work} for baselines) on both LLama 8B and GPT-4 with code interpreter enabled. Notably, while GPT-4 can potentially generate and execute solver code, our ACCORD-based method enables the LLM to generate solutions end-to-end without code execution. For both our approach and all prompting baselines, 60 samples per instance are generated, and the best result is selected. ACCORD consistently outperforms prompting strategies across all 6 combinatorial optimization tasks, and achieves optimal solutions on smaller instances.
We also assess the impact of model size on average gap, feasibility, and inference time (Fig.~\ref{fig:model_size_effect}). The 8B model mostly outperforms the 1B model in feasibility and optimality gap, with only a moderate increase in inference time. For harder instances, such as JSSP, the 1B model fails to find feasible solutions within the sampling limit. Our results demonstrate that scaling from 1B to 8B parameters yields a significant 31.5\% relative improvement in solution quality, reducing the average gap from 6.54\% to 4.48\% (Table \ref{tab:gap_comparison_small_table}). The most substantial improvements were observed in routing problems, with TSP and VRP showing 65\% and 54\% relative gap reductions, respectively. Bin packing problems showed minimal sensitivity to model scale, with only a 1\% improvement.
In addition to our synthetic OR-Tools instances, we also evaluated ACCORD-8B on Taillard permutation flow-shop benchmarks (50 jobs × 10 machines and 50 jobs × 20 machines; avg.\ gap $\approx 13.7 \%$) and on job-shop benchmarks TAI\cite{taillard1993benchmarks} (15 × 15 to 50 × 20; avg.\ gap $\approx 21.7 \%$) and DMU\cite{dmu_dataset} (20 × 15 to 50 × 15; avg.\ gap $\approx 22.1 \%$) against standard heuristics (MWR/MOR/SPT) and the L2D neural scheduler (see Supplementary Material for full results and runtimes).

\subsection{Relationship Between Latent Space Proximity and Solution Feasibility}
\label{sec:stat_test}
\begin{figure}[h]
    \centering
    \includegraphics[width=0.8\linewidth]{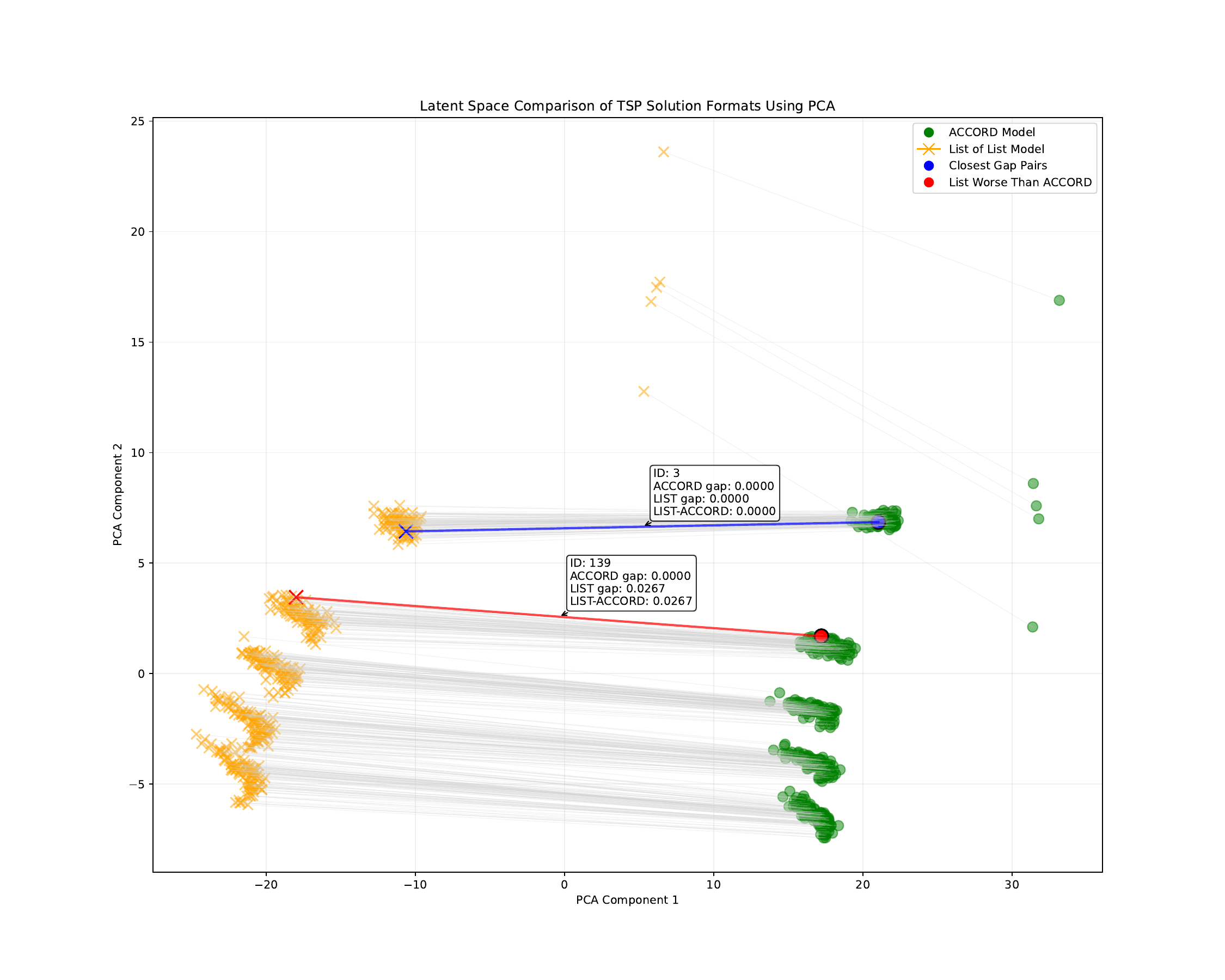}
    \caption{Latent representation distance versus solution feasibility on TSP problems, demonstrating negative correlation between distance and constraint satisfaction.}
    \label{fig:lat_dist_feasibility}
\end{figure}

To investigate the connection between latent representations and solution feasibility, we analyzed 500 TSP instances processed using both ACCORD and list-of-list formats. For each instance, we extracted hidden-state representations from the final transformer layer of the LLaMA 3.1 8B model with PCA deimentionality reduction, then computed the Euclidean distance between paired representations from each format. We subsequently evaluated the feasibility of the solutions generated by both models. Statistical analysis revealed a significant negative correlation between latent distance and solution feasibility ($r = -0.1082$, $p = 0.0155$, $p < 0.05$), indicating that solutions whose latent representations are closer to those produced by the ACCORD format are more likely to satisfy constraints(see Figure \ref{fig:lat_dist_feasibility}). This trend was further supported by quartile analysis, which showed feasibility rates consistently decreasing as latent distance increased. Notably, this relationship holds despite a large performance gap between the formats (71.4\% feasible solutions for ACCORD vs.\ 1.6\% for list-of-lists).
These findings suggest that LLMs encode constraint satisfaction geometrically: solutions closer to the ACCORD manifold in latent space are more likely to be feasible. Thus, latent proximity can predict solution quality, indicating that neural solvers capture structural information about combinatorial constraints beyond explicit training signals.

\section{Conclusion}
We introduced ACCORD, a framework that encodes combinatorial constraints into an autoregressive text format and uses dynamic LoRA routing to probe an LLM’s end-to-end ability on NP-hard tasks. On six standard benchmarks (TSP, VRP, FlowShop, JSSP, Knapsack, BinPacking), an 8 B-parameter model trained with ACCORD achieves strong feasibility rates and competitive optimality gaps compared to prompting and a naïve list-of-lists format. Our goal is not to supplant specialized solvers but to map out how far small LLMs can go as self-contained combinatorial reasoners. By releasing ACCORD and its 90K dataset, we offer a reproducible codebase for future work at the intersection of optimization and generative modeling.
\section{Limitations and Future Work}
\label{sec:limits}
Despite its strong performance, ACCORD is bounded by the LLM’s context window (limiting very large instances) and relies on LoRA adapters on an 8B-parameter model. In future work, we will investigate larger backbones (with full fine-tuning), expand the effective context via external memory or hierarchical encoding, and apply ACCORD to real-world, large-scale optimization scenarios.

\clearpage


\bibliographystyle{plain}
\bibliography{main}

\begin{thebibliography}{10}

\bibitem{abgaryan2024starjob}
H.~Abgaryan, T.~Cazenave, and A.~Harutyunyan.
\newblock Starjob: Dataset for llm-driven job shop scheduling.
\newblock In {\em ArXiv Preprint arXiv:2503.01877v1}, 2024.

\bibitem{adams1988shifting}
James Adams, Elias Balas, and David Zawack.
\newblock Shifting bottleneck procedures for job shop scheduling.
\newblock In {\em Management Science}, volume~34, pages 391--401. INFORMS, 1988.

\bibitem{llama3modelcard}
Meta AI.
\newblock Llama 3 model card, 2024.
\newblock Accessed: 2024-08-10.

\bibitem{unslothai_unsloth}
Unsloth AI.
\newblock Unsloth: Accelerated fine-tuning for large language models, 2024.
\newblock Accessed: 2024-11-19.

\bibitem{Awasthi2025}
Pranjal Awasthi, Sreenivas Gollapudi, Ravi Kumar, and Kamesh Munagala.
\newblock Combinatorial optimization via llm-driven iterated fine-tuning.
\newblock {\em arXiv preprint arXiv:2503.06917}, 2025.

\bibitem{brown2020language}
Tom Brown, Benjamin Mann, Nick Ryder, Melanie Subbiah, Jared~D Kaplan, Prafulla Dhariwal, Arvind Neelakantan, Pranav Shyam, Girish Sastry, Amanda Askell, et~al.
\newblock Language models are few-shot learners.
\newblock {\em Advances in neural information processing systems}, 33:1877--1901, 2020.

\bibitem{chowdhery2023palm}
Aakanksha Chowdhery, Sharan Narang, Jacob Devlin, Maarten Bosma, Gaurav Mishra, Adam Roberts, Paul Barham, Hyung~Won Chung, Charles Sutton, Sebastian Gehrmann, et~al.
\newblock Palm: Scaling language modeling with pathways.
\newblock {\em Journal of Machine Learning Research}, 24(240):1--113, 2023.

\bibitem{dmu_dataset}
Ebru Demirkol, Sanjay Mehta, and Reha Uzsoy.
\newblock Benchmarks for shop scheduling problems.
\newblock {\em European Journal of Operational Research}, 109(1):137--141, 1998.

\bibitem{goel1996genetic}
M~Goel et~al.
\newblock Genetic algorithms in scheduling.
\newblock In {\em International Conference on Genetic Algorithms}, 1996.

\bibitem{ortools}
{Google}.
\newblock Google's or-tools.
\newblock \url{https://developers.google.com/optimization/}.
\newblock Accessed: 2024-05-07.

\bibitem{hu2022lora}
Edward~J Hu, yelong shen, Phillip Wallis, Zeyuan Allen-Zhu, Yuanzhi Li, Shean Wang, Lu~Wang, and Weizhu Chen.
\newblock Lo{RA}: Low-rank adaptation of large language models.
\newblock In {\em International Conference on Learning Representations}, 2022.

\bibitem{huang2022improving}
W.~Huang, P.~Abbeel, D.~Pathak, and I.~Mordatch.
\newblock Improving the reasoning capabilities of large language models in complex tasks.
\newblock In {\em Proceedings of the International Conference on Machine Learning}, 2022.

\bibitem{huang2024large}
W.~Huang et~al.
\newblock Large language models for vehicle routing: A prompting-based approach.
\newblock In {\em Proceedings of the Conference on Empirical Methods in Natural Language Processing}, 2024.

\bibitem{SGE}
Zangir Iklassov, Yali Du, Farkhad Akimov, and Martin Tak\'{a}\v{c}.
\newblock Self-guiding exploration for combinatorial problems.
\newblock In {\em Advances in Neural Information Processing Systems 37 (NeurIPS 2024)}, 2024.

\bibitem{kalajdzievski2023rankstabilizationscalingfactor}
Damjan Kalajdzievski.
\newblock A rank stabilization scaling factor for fine-tuning with lora, 2023.

\bibitem{khalil2017learning}
Elias~B. Khalil, Hanjun Dai, Yuyu Zhang, Bistra Dilkina, and Le~Song.
\newblock Learning combinatorial optimization algorithms over graphs.
\newblock In {\em Advances in Neural Information Processing Systems}, 2017.

\bibitem{kool2019attention}
Wouter Kool, Holger van Hoof, and Max Welling.
\newblock Attention, learn to solve routing problems!
\newblock In {\em International Conference on Learning Representations}, 2019.

\bibitem{lenstra1979complexity}
Jan~K Lenstra, A~Rinnooy~Kan, and P~Brucker.
\newblock Complexity of machine scheduling problems.
\newblock {\em Annals of Discrete Mathematics}, 1:343--362, 1979.

\bibitem{Liu2023}
Shengcai Liu, Caishun Chen, Xinghua Qu, Ke~Tang, and Yew-Soon Ong.
\newblock Large language models as evolutionary optimizers.
\newblock {\em arXiv preprint arXiv:2310.19046}, 2023.

\bibitem{madaan2023self}
Aman Madaan, Niket Tandon, Prakhar Gupta, Skyler Hallinan, Luyu Gao, Sarah Wiegreffe, Uri Alon, Nouha Dziri, Shrimai Prabhumoye, Yiming Yang, et~al.
\newblock Self-refine: Iterative refinement with self-feedback.
\newblock {\em Advances in Neural Information Processing Systems}, 36:46534--46594, 2023.

\bibitem{Masoud2024}
Mahmoud Masoud, Ahmed Abdelhay, and Mohammed Elhenawy.
\newblock Exploring combinatorial problem solving with large language models: A case study on the traveling salesman problem using gpt-3.5 turbo.
\newblock {\em arXiv preprint arXiv:2405.01997}, 2024.

\bibitem{Mittal2024}
Chinmay Mittal, Krishna Kartik, Mausam, and Parag Singla.
\newblock Puzzlebench: Can llms solve challenging first-order combinatorial reasoning problems?
\newblock {\em arXiv preprint arXiv:2402.02611}, 2024.

\bibitem{nawaz1983heuristic}
Muhammad Nawaz, E.~Emory Enscore, and Inyong Ham.
\newblock A heuristic algorithm for the m-machine, n-job flow-shop sequencing problem.
\newblock {\em Omega}, 11(1):91--95, 1983.

\bibitem{oroojlooyjadid2020applying}
Afshin Oroojlooyjadid, Lawrence~V Snyder, and Martin Tak{\'a}{\v{c}}.
\newblock Applying deep learning to the newsvendor problem.
\newblock {\em Iise Transactions}, 52(4):444--463, 2020.

\bibitem{roy1964machine}
R~Roy and G~Sussmann.
\newblock Machine scheduling by mathematical programming.
\newblock In {\em Journal of the Operational Research Society}, volume~15, pages 352--362. JORS, 1964.

\bibitem{taillard1993benchmarks}
Eric Taillard.
\newblock Benchmarks for basic scheduling problems.
\newblock {\em European Journal of Operational Research}, 64(2):278--285, 1993.

\bibitem{thoppilan2022lamda}
Romal Thoppilan, Daniel De~Freitas, Jamie Hall, Noam Shazeer, Apoorv Kulshreshtha, Heng-Tze Cheng, Alicia Jin, Taylor Bos, Leslie Baker, Yu~Du, et~al.
\newblock Lamda: Language models for dialog applications.
\newblock {\em arXiv preprint arXiv:2201.08239}, 2022.

\bibitem{valmeekam2022benchmark}
K.~Valmeekam, A.~Olmo, S.~Sreedharan, and S.~Kambhampati.
\newblock A benchmark for evaluating planning and reasoning in large language models.
\newblock In {\em NeurIPS Foundation Models for Decision Making Workshop}, 2022.

\bibitem{Wan2024}
Fang Wan, Julien Fondrevelle, Tao Wang, Kezhi Wang, and Antoine Duclos.
\newblock Optimizing small-scale surgery scheduling with large language model.
\newblock In {\em Proceedings of the 21st International Conference on Informatics in Control, Automation and Robotics (ICINCO)}, pages 223--228, Lisbon, Portugal, 2024.

\bibitem{wei2022emergent}
Jason Wei, Yi~Tay, Rishi Bommasani, Colin Raffel, Barret Zoph, Sebastian Borgeaud, Dani Yogatama, Maarten Bosma, Denny Zhou, Donald Metzler, et~al.
\newblock Emergent abilities of large language models.
\newblock {\em arXiv preprint arXiv:2206.07682}, 2022.

\bibitem{wei2022chain}
Jason Wei, Xuezhi Wang, Dale Schuurmans, Maarten Bosma, Fei Xia, Ed~Chi, Quoc~V Le, Denny Zhou, et~al.
\newblock Chain-of-thought prompting elicits reasoning in large language models.
\newblock {\em Advances in neural information processing systems}, 35:24824--24837, 2022.

\bibitem{yang2023optimization}
C.~Yang, X.~Wang, Y.~Lu, H.~Liu, Q.V. Le, and X.~Chen.
\newblock Optimization by prompting: Leveraging large language models for combinatorial optimization.
\newblock {\em arXiv preprint arXiv:2309.03409}, 2023.

\bibitem{zhang2020learning}
C.~Zhang, W.~Song, Z.~Cao, J.~Zhang, P.S. Tan, and C.~Xu.
\newblock Learning to dispatch for job shop scheduling via deep reinforcement learning.
\newblock In {\em Advances in Neural Information Processing Systems}, 2020.

\bibitem{zhang2022automatic}
Zhuosheng Zhang, Aston Zhang, Mu~Li, and Alex Smola.
\newblock Automatic chain of thought prompting in large language models.
\newblock {\em arXiv preprint arXiv:2210.03493}, 2022.

\bibitem{zhou2022least}
Denny Zhou, Nathanael Sch{\"a}rli, Le~Hou, Jason Wei, Nathan Scales, Xuezhi Wang, Dale Schuurmans, Claire Cui, Olivier Bousquet, Quoc Le, et~al.
\newblock Least-to-most prompting enables complex reasoning in large language models.
\newblock {\em arXiv preprint arXiv:2205.10625}, 2022.

\end{thebibliography}
\clearpage
\appendix

\section{Technical Appendices and Supplementary Material}

\appendix

\appendix
\section*{Appendix: Example Combinatorial Optimization Problems}
\label{app:example_solution}
Below, we provide representative examples from each of the major combinatorial optimization problems in our dataset. Each example is shown in both the standard ``list of list'' representation and the ``Accord'' (arrow-based) representation.

\subsection*{Knapsack Problem}

\textbf{Instruction:} You are given a paired representation (value, weight): Find a set of items to pack into a container with a maximum weight capacity $= 20$ that maximizes total value of packed items.

\begin{itemize}
    \item \textbf{Input:}
    \[
    \begin{bmatrix}
    6 & 5 \\
    10 & 8 \\
    7 & 4 \\
    6 & 2 \\
    1 & 1 \\
    \end{bmatrix}
    \]
    \item \textbf{Output (List of List):}
    \begin{verbatim}
Solution: [(6, 5), (10, 8), (7, 4), (6, 2), (1, 1)]
  Value: 6+10+7+6+1=30
  Weight: 5+8+4+2+1=20<=20
    \end{verbatim}
    \item \textbf{Output (Accord):}
    \begin{verbatim}
Solution:
[[6, 5] -> value:0+6=6, weight:0+5=5<=20],
[[10, 8] -> value:6+10=16, weight:5+8=13<=20],
[[7, 4] -> value:16+7=23, weight:13+4=17<=20],
[[6, 2] -> value:23+6=29, weight:17+2=19<=20],
[[1, 1] -> value:29+1=30, weight:19+1=20<=20]

Total Value: 30
Total Weight: 20<=20
    \end{verbatim}
\end{itemize}

\subsection*{Bin Packing Problem}

\textbf{Instruction:} Given a list of items (id, weight), determine the minimum number of bins (capacity=77) needed to pack all items without exceeding the capacity.

\begin{itemize}
    \item \textbf{Input:}
    \[
    [(0,17), (1,24), (2,19), (3,11), (4,27)]
    \]
    \item \textbf{Output (List of List):}
    \begin{verbatim}
The minimum number of bins required is 2. The bin assignments are: [[0, 1, 2, 3], [4]].
    \end{verbatim}
    \item \textbf{Output (Accord):}
    \begin{verbatim}
Bin 1:
(0, 17)->17 (1, 24)->41 (2, 19)->60 (3, 11)->71<=77
Bin 2:
(4, 27)->27<=77
Total bins required: 2
    \end{verbatim}
\end{itemize}

\subsection*{Vehicle Routing Problem (VRP)}

\textbf{Instruction:} Given customers with coordinates and a depot, and multiple vehicles of capacity 85, find the minimum-length routes serving all customers.

\begin{itemize}
    \item \textbf{Coords:}
    0:(34, 42), 1:(39, 58), 2:(46, 48), 3:(57, 49), 4:(45, 16)
    \item \textbf{Output (List of List):}
    \begin{verbatim}
[(0): (34, 42), (0): (34, 42)]
[(0): (34, 42), (0): (34, 42)]
[(0): (34, 42), (0): (34, 42)]
[(0): (34, 42), (0): (34, 42)]
[(0): (34, 42), (1): (39, 58), (2): (46, 48), 
(3): (57, 49), (4): (45, 16), (0): (34, 42)]
Overall Total Distance: 102
    \end{verbatim}
    \item \textbf{Output (Accord):}
    \begin{verbatim}
Vehicle Route: (0): (34, 42) -> (0): (34, 42) + 0
Vehicle Route: (0): (34, 42) -> (0): (34, 42) + 0
Vehicle Route: (0): (34, 42) -> (0): (34, 42) + 0
Vehicle Route: (0): (34, 42) -> (0): (34, 42) + 0
Vehicle Route: (0): (34, 42) -> (1): (39, 58) + 16 -> (2): (46, 48) + 12 -> 
(3): (57, 49) + 11 -> (4): (45, 16) + 35 -> (0): (34, 42) + 28
Overall Total Distance: 102
    \end{verbatim}
\end{itemize}

\subsection*{Traveling Salesman Problem (TSP)}

\textbf{Instruction:} Given customers with coordinates and a depot, and 1 vehicle, find the minimum-length route serving all customers.

\begin{itemize}
    \item \textbf{Coords:}
    0:(17, 22), 1:(63, 8), 2:(22, 60), 3:(3, 29), 4:(7, 12)
    \item \textbf{Output (List of List):}
    \begin{verbatim}
[(0): (17, 22), (4): (7, 12), (3): (3, 29), (2): (22, 60), (1): (63, 8), (0): (17, 22)]
Overall Total Distance: 181
    \end{verbatim}
    \item \textbf{Output (Accord):}
    \begin{verbatim}
Vehicle Route: (0): (17, 22) -> (4): (7, 12) + 14 -> (3): (3, 29) + 17 -> 
(2): (22, 60) + 36 -> (1): (63, 8) + 66 -> (0): (17, 22) + 48
Overall Total Distance: 181
    \end{verbatim}
\end{itemize}

\subsection*{Job Shop Scheduling Problem (JSSP)}

\textbf{Instruction:} Optimize schedule for 2 Jobs (J) across 6 Machines (M) to minimize makespan. Each M can process only one J at a time, and once started, J cannot be interrupted.

\begin{itemize}
    \item \textbf{Input:}
    \begin{verbatim}
J0:
M2:205 M1:157 M0:198 M5:79 M3:110 M4:32 
J1:
M3:179 M4:108 M2:82 M5:112 M1:136 M0:27 
    \end{verbatim}
    \item \textbf{Output (List of List):}
    \begin{verbatim}
[[0, 2, 0, 205], [1, 3, 0, 179], [1, 4, 179, 108], [0, 1, 205, 157], [1, 2, 287, 82], 
[0, 0, 362, 198], [1, 5, 369, 112], [1, 1, 481, 136], [0, 5, 560, 79], [1, 0, 617, 27], 
[0, 3, 639, 110], [0, 4, 749, 32]]
Maximum end completion time or Makespan: 781
    \end{verbatim}
    \item \textbf{Output (Accord):}
    \begin{verbatim}
Solution:
J0-M2: 0+205 -> 205,
J1-M3: 0+179 -> 179,
J1-M4: 179+108 -> 287,
J0-M1: 205+157 -> 362,
J1-M2: 287+82 -> 369,
J0-M0: 362+198 -> 560,
J1-M5: 369+112 -> 481,
J1-M1: 481+136 -> 617,
J0-M5: 560+79 -> 639,
J1-M0: 617+27 -> 644,
J0-M3: 639+110 -> 749,
J0-M4: 749+32 -> 781,
Maximum end completion time or Makespan: 781 
    \end{verbatim}
\end{itemize}

\subsection*{Flow Shop Scheduling Problem (FSSP)}

\textbf{Input:}
\begin{verbatim}
J1:
M1:12 M2:7 
J2:
M1:8 M2:4 
J3:
M1:4 M2:15 
J4:
M1:5 M2:9 
\end{verbatim}
\textbf{Output (List of List):}
\begin{verbatim}
[[3, 1, 0, 4], [3, 2, 4, 15], [2, 1, 4, 8], [4, 1, 12, 5], 
[1, 1, 17, 12], [2, 2, 19, 4], [4, 2, 23, 9], [1, 2, 32, 7]]
Maximum end completion time or Makespan: 39
\end{verbatim}
\textbf{Output (Accord):}
\begin{verbatim}
J3: M1(0+4=4) -> M2(4+15=19)
J2: M1(4+8=12) -> M2(19+4=23)
J4: M1(12+5=17) -> M2(23+9=32)
J1: M1(17+12=29) -> M2(32+7=39)

Maximum end completion time or Makespan: 39
\end{verbatim}


\section{TextClassifier Routing model archirtecture with Dynamic attention}
\label{app:in_detail_routing_arch}

In order to activate correct LoRA layers corresponding to each combinatorial optimization problem being solved, we utilize an Attention based
Dynamic Router TextClassifier which dynamically activates the appropriate LoRA weights based on the instruction text input. The complete pipeline is presented in Figure~\ref{fig:pipline}.

Our model builds upon transformer-based architectures with several key enhancements to effectively capture problem-specific features. Given an input sequence of tokens $\mathbf{x} = (x_1, x_2, \ldots, x_n)$ where each $x_i$ represents a token from vocabulary $\mathcal{V}$, we first map each token to a $d_e$-dimensional embedding space. The embedding layer combines token embeddings with positional information:

\begin{equation}
    \mathbf{E} = \mathbf{E}_{\text{token}}(\mathbf{x}) + \mathbf{E}_{\text{pos}}(\mathbf{p})
\end{equation}

Here, $\mathbf{E}_{\text{token}} \in \mathbb{R}^{|\mathcal{V}| \times d_e}$ is the token embedding matrix, $\mathbf{E}_{\text{pos}} \in \mathbb{R}^{n_{\max} \times d_e}$ is the positional embedding matrix (where $n_{\max}$ is the maximum sequence length), and $\mathbf{p} = (1, 2, \ldots, n)$ are the position indices. To enhance representation stability, we apply layer normalization and dropout:

\begin{equation}
    \mathbf{E}' = \text{Dropout}(\text{LayerNorm}(\mathbf{E}))
\end{equation}

The embeddings are then projected to a hidden dimension $d_h$ through a linear transformation:
\begin{equation}
    \mathbf{H}_0 = \mathbf{E}'\mathbf{W}_p + \mathbf{b}_p
\end{equation}
where $\mathbf{W}_p \in \mathbb{R}^{d_e \times d_h}$ and $\mathbf{b}_p \in \mathbb{R}^{d_h}$ are learnable parameters.

The projected embeddings $\mathbf{H}_0$ are processed through multiple transformer layers, where each layer $l \in \{1,2,3\}$ applies multi-head attention followed by normalization and feed-forward processing:

\begin{align}
    \mathbf{H}_l = 
    \begin{cases}
        \text{LayerNorm}(\mathbf{H}_{l-1} + \text{MultiHead}(\mathbf{H}_{l-1})), & \text{if } l \in \{1,3\} \\
        \text{LayerNorm}(\mathbf{H}_{l-1} + \text{FFN}(\mathbf{H}_{l-1})), & \text{if } l = 2
    \end{cases}
\end{align}

The sequence of token representations in $\mathbf{H}_3$ is converted into a single fixed-length vector using an attention-based pooling mechanism that learns to assign importance weights to different tokens:

\begin{equation}
    \mathbf{r} = \sum_{i=1}^{n} a_i \mathbf{h}_{3,i}
\end{equation}

Finally, the pooled representation $\mathbf{r}$ is passed through a classification head with learnable parameters $\mathbf{W}_1$, $\mathbf{b}_1$, $\mathbf{W}_2$, and $\mathbf{b}_2$:

\begin{equation}
    \mathbf{y} = \mathbf{W}_2 \cdot \text{LayerNorm}(\text{GELU}(\mathbf{W}_1 \mathbf{r} + \mathbf{b}_1)) + \mathbf{b}_2
\end{equation}

The output $\mathbf{y} \in \mathbb{R}^c$ represents the logits for each of the $c$ combinatorial optimization problem classes.

\subsection{Dataset Generation Details}
\label{sec:dataset_appendix}

We generated synthetic supervised datasets for several CO problems using Google OR-Tools \cite{ortools} as the primary solver. For each problem instance, we generated solutions in two formats: the conventional ``list of lists'' representation (see Appendix \ref{app:example_solution}) and our proposed ACCORD representation (see Section \ref{sec:accord} and Appendix \ref{app:example_solution}). Approximately 15,000 instances with corresponding solutions in both formats were generated for each problem category.

\textbf{TSP \& VRP}
A combined dataset was generated for the Traveling Salesperson Problem (TSP) and Vehicle Routing Problem (VRP). Instances varied by the number of locations $N \in \{5, 8, 10, 12, 15, 20, 50, 75, 100\}$ and the number of vehicles $V \in \{1, \dots, 10\}$. TSP instances used $V=1$, while VRP used $V>1$. Locations had random integer coordinates, and demands were assigned randomly (depot demand $d_0=0$). Vehicle capacity constraints were included for VRP. The objective was minimizing total Euclidean distance. Google OR-Tools solved instances using the `PATH\_CHEAPEST\_ARC` strategy.
\textbf{Knapsack}
Instances of the 0/1 Knapsack Problem were generated with varying item counts $N \in \{5, 8, 10, 12, 15, 20, 25, 30, 50, 100\}$ and categorized by difficulty ("easy", "medium", "hard"). Difficulty influenced item value/weight ranges, the ratio of total item weight to capacity, and value-weight correlations. Optimal solutions (maximizing value within capacity) were computed using OR-Tools' `KNAPSACK\_MULTIDIMENSION\_BRANCH\_AND\_BOUND\_SOLVER` with a 180s timeout per instance; timed-out instances were discarded.
\textbf{Bin Packing}
Instances for the Bin Packing Problem varied by item counts $N \in \{5, 8, 12, 15, 20, 50, 100\}$, item weight ranges (maximums of 10, 20, 50, 100), and target solution bins $B \in \{1, \dots, 10\}$. Bin capacity was determined based on total item weight and the target bin count. OR-Tools found optimal bin assignments, minimizing the number of bins used, subject to a 180s timeout. The generation aimed for a balanced distribution across target bin counts.
\textbf{JSSP}
Instances for the Job Shop Scheduling Problem (JSSP) were generated for various dimensions (jobs $\times$ machines), including configurations like 10x10, 20x20, 50x20, 100x20, etc. Machine sequences for jobs were random permutations, and operation durations were random integers (range 5-300). The objective was makespan minimization. Solutions were found using the OR-Tools CP-SAT solver with an 8-worker parallel search and a 3600s timeout.
\textbf{FSSP}
A dataset for the Permutation Flowshop Scheduling Problem (PFSP) was generated with dimensions (jobs $\times$ machines) ranging from 5x1 to 50x2 and 2x50. Processing times were random integers (range 1-100). The objective was to find a single job permutation minimizing makespan. For this dataset, solutions were generated using the NEH heuristic \cite{nawaz1983heuristic}.

\section*{Training details}
\label{app:train_details}
The model being fine-tuned is LLaMA 3.1, an 8 billion parameter model from Meta\cite{llama3modelcard}, using a 4-bit quantized version to reduce memory usage. Finetning was conducted using Stabilized Low-Rank Adaptation (RsLoRA) \cite{kalajdzievski2023rankstabilizationscalingfactor} with rank \(r = 64\) to introduce learnable parameters specifically in targeted layers. \cite{kalajdzievski2023rankstabilizationscalingfactor} Compared to Lora\cite{hu2022lora} RsLoRa improves the stability of training by modifying the rank during adaptation\cite{kalajdzievski2023rankstabilizationscalingfactor}. 
The target modules include:
\begin{multline}
\text{target\_modules} = \{\texttt{q\_proj},\ \texttt{k\_proj},\ \texttt{v\_proj},\ \texttt{o\_proj},\\
\texttt{gate\_proj},\ \texttt{up\_proj},\ \texttt{down\_proj}\}
\end{multline}
The LoRA-specific parameters are configured as follows:
\begin{itemize}
    \item Rank (\(r\)): 64
    \item LoRA Alpha (\(\alpha\)): 64
    \item LoRA Dropout: 0
    \item Bias: none
\end{itemize}
This resulted in number of trainable parameters $= 167,772,160$ or 0.02 \% of the entire Llama 8B model's parameters.

\section*{Quantization and Memory Efficiency}
The model is loaded in 4-bit precision to reduce memory consumption. Gradient checkpointing is enabled using the \texttt{unsloth} \cite{unslothai_unsloth} method, to fit longer sequences by saving memory. This reduces the VRAM usage by approximately 30\%, enabling larger batch sizes.

\begin{figure}[htbp]
    \centering
    \begin{subfigure}[t]{0.45\textwidth}
        \centering
        \includegraphics[width=\linewidth]{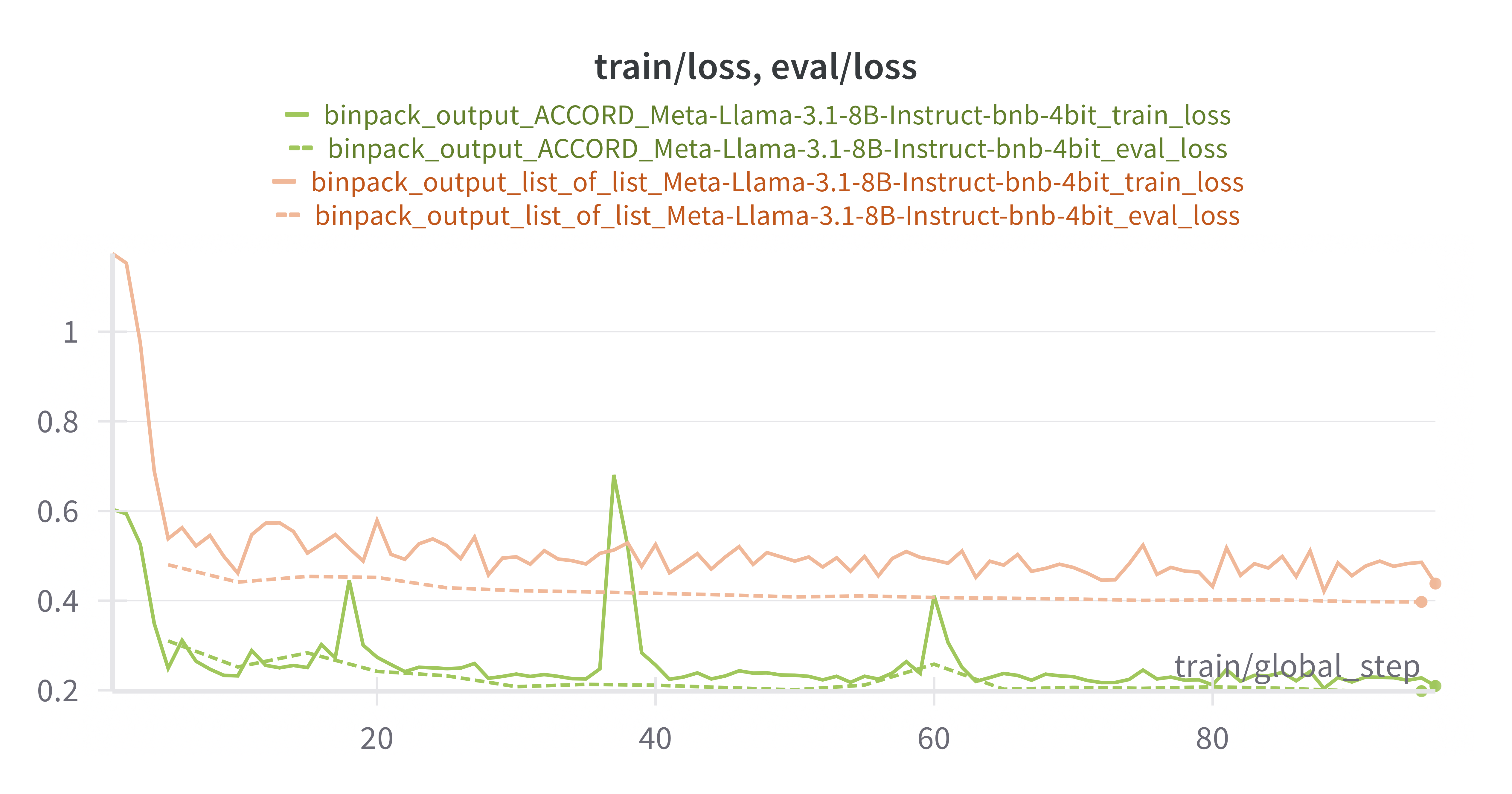}
        \caption{BinPack train and validation loss}
    \end{subfigure}\hfill
    \begin{subfigure}[t]{0.45\textwidth}
        \centering
        \includegraphics[width=\linewidth]{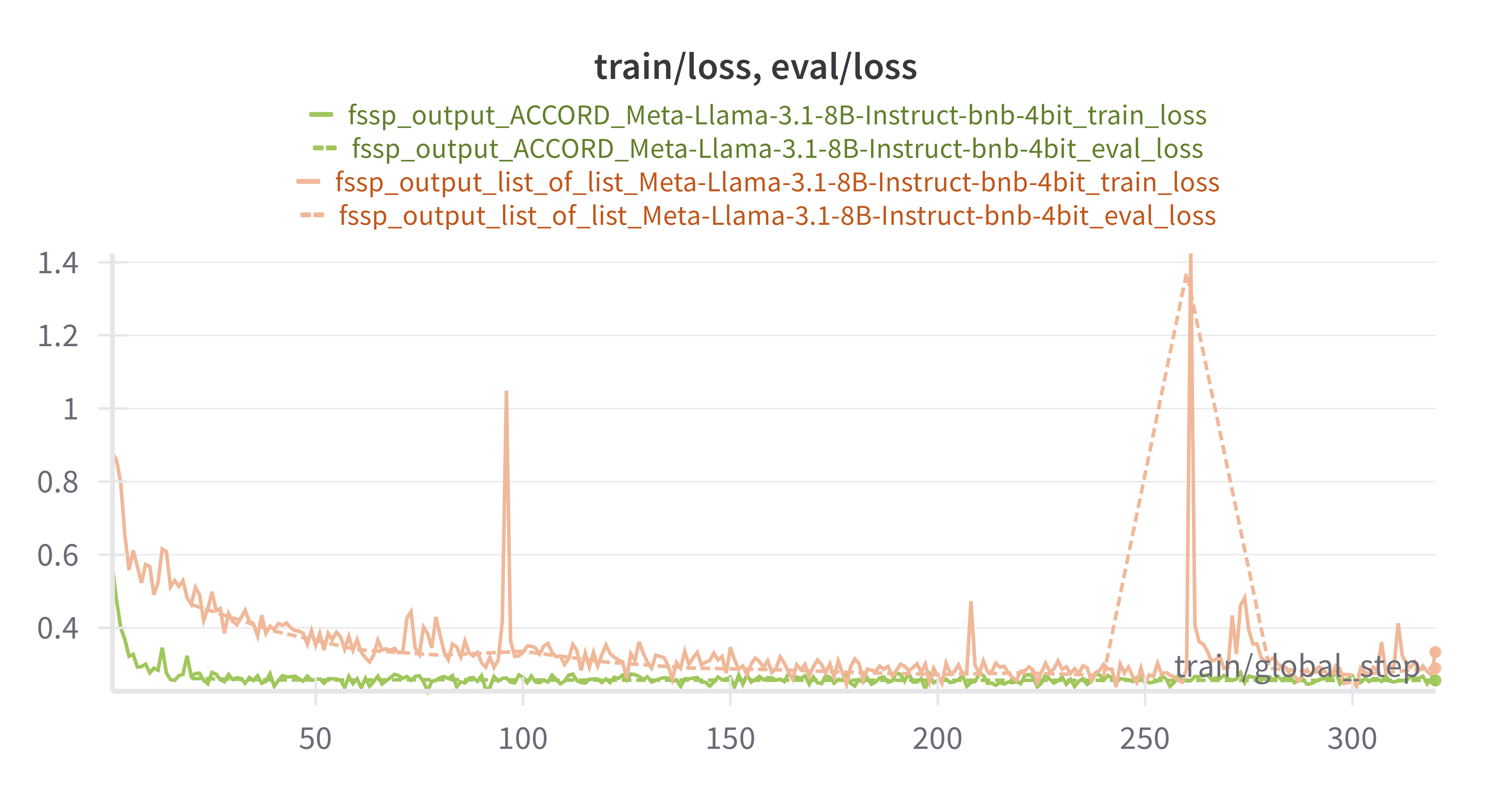}
        \caption{FlowShop train and validation loss}
    \end{subfigure}

    \vspace{1em} 

    \begin{subfigure}[t]{0.45\textwidth}
        \centering
        \includegraphics[width=\linewidth]{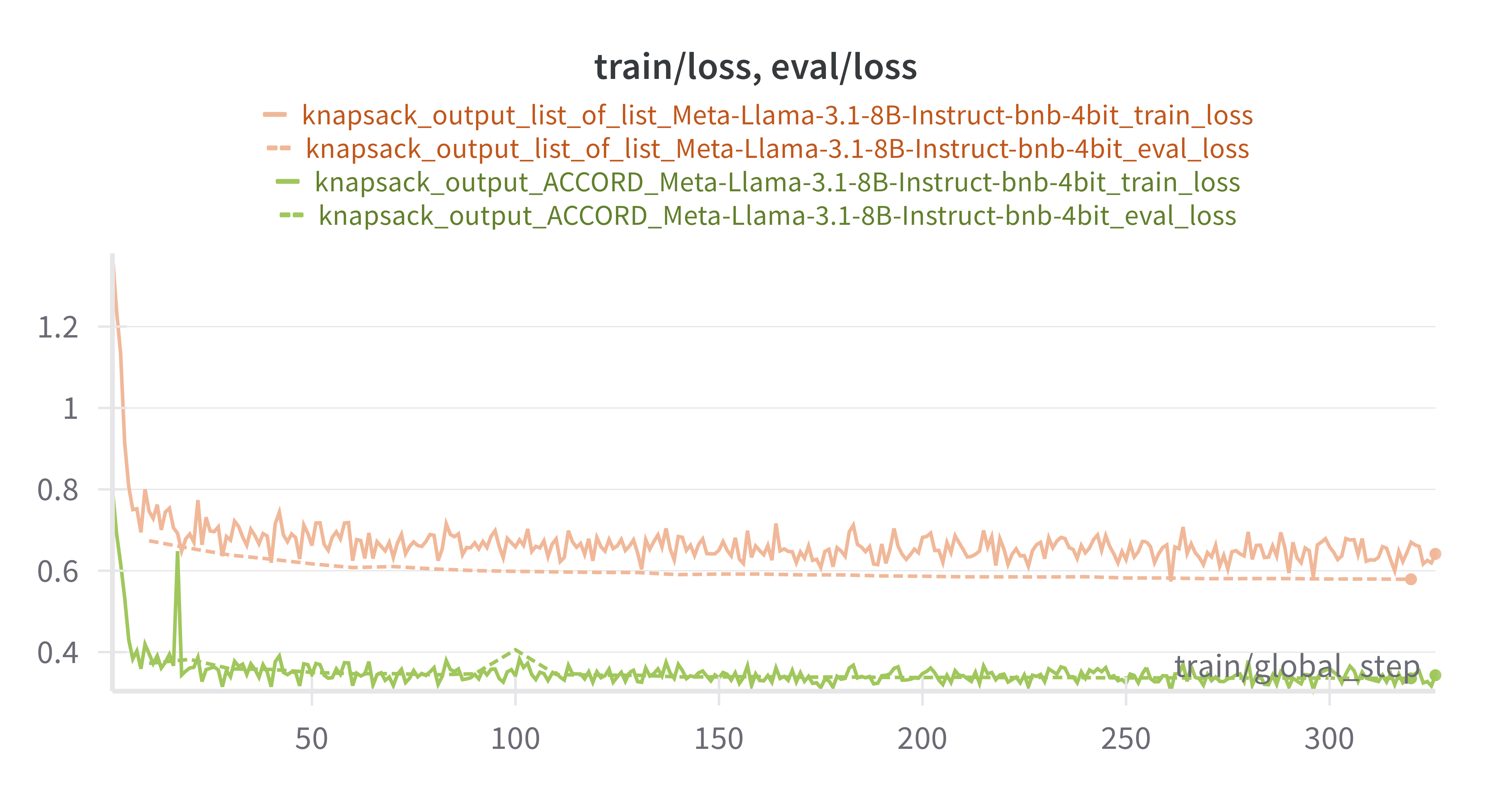}
        \caption{Knapsack train and validation loss}
    \end{subfigure}\hfill
    \begin{subfigure}[t]{0.45\textwidth}
        \centering
        \includegraphics[width=\linewidth]{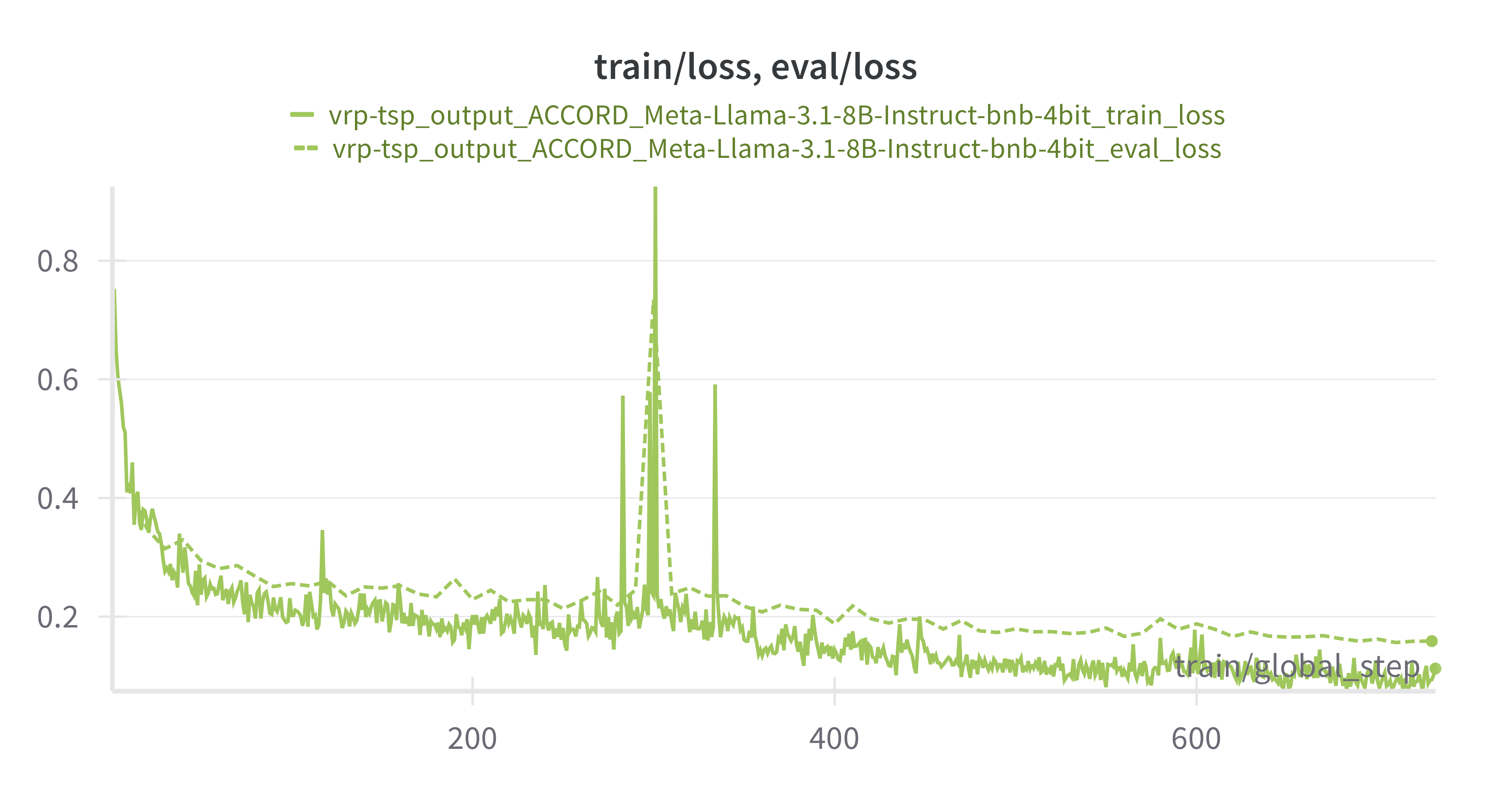}
        \caption{VRP-TSP train and validation loss}
    \end{subfigure}

    \vspace{1em} 

    \begin{subfigure}[t]{0.45\textwidth}
        \centering
        \includegraphics[width=\linewidth]{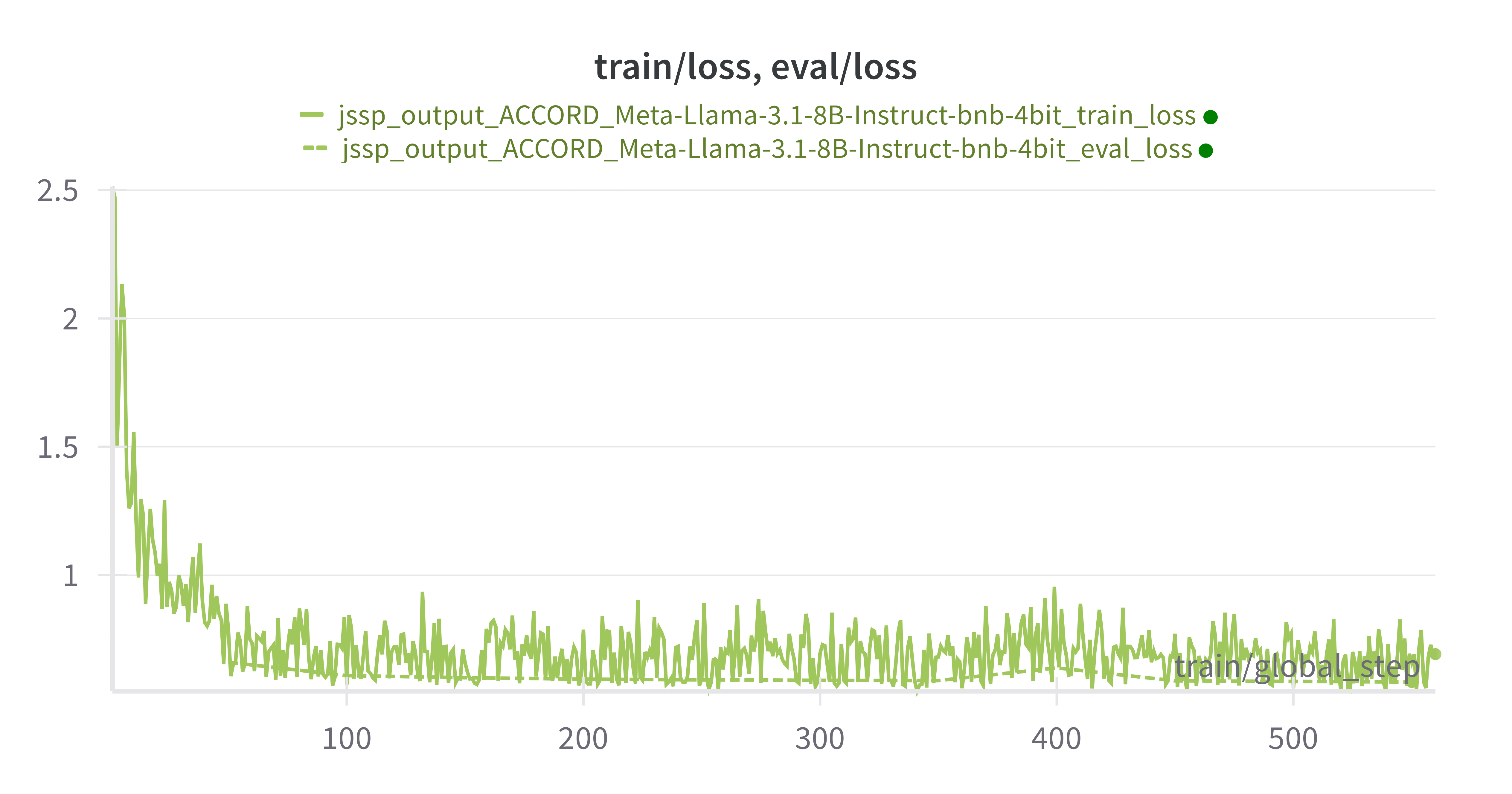}
        \caption{JSSP train and validation loss}
    \end{subfigure}\hfill
    \caption{Training and evaluation losses of Llama 3.1 8B model on ACCORD dataset for Various tasks. Finetuning has been done using lora rank 64 and lora sacle 64 hyperparameters.}
    \label{fig:loss_plots}
\end{figure}

\section*{Training Parameters}
The fine-tuning process is controlled by the following parameters:
\begin{itemize}
    \item \textbf{Batch size per device}: 8
    \item \textbf{Gradient accumulation steps}: 8
    \item \textbf{Max sequence length}: 10,000 tokens
    \item \textbf{Number of epochs}: 1
    \item \textbf{Warmup steps}: 5
    \item \textbf{Learning rate}: \(2 \times 10^{-4}\)
    \item \textbf{Optimizer}: AdamW with 8-bit precision
    \item \textbf{Weight decay}: 0.01
    \item \textbf{Learning rate scheduler}: Linear decay
    \item \textbf{FP16 precision}:True
    \item \textbf{Number of Epochs}: 1
\end{itemize}
\begin{table}[ht]
\centering
\caption{The effect of the model size on Average Gap (\%): Comparison Across CO Problems }
\label{tab:gap_comparison_small_table}
\begin{tabular}{lcc}
\hline
\textbf{Problem} & \textbf{1B Model} & \textbf{8B Model} \\
\hline
BINPACK & 1.01\% & 1.00\% \\
FSSP & 7.92\% & 7.17\% \\
JSSP & N/A & 6.08\% \\
KNAPSAK & 5.90\% & 5.33\% \\
TSP & 8.11\% & 2.84\% \\
VRP & 9.74\% & 4.50\% \\
\hline
\textbf{AVERAGE} & 6.54\% & 4.48\% \\
\hline
\end{tabular}
\end{table}

\end{document}